\documentclass{article}

\usepackage{arxiv}

\usepackage[utf8]{inputenc} 
\usepackage[T1]{fontenc}    
\usepackage{color,xcolor}
\usepackage{wrapfig}
\usepackage[pagebackref=true,colorlinks,citecolor=brown]{hyperref}
\renewcommand{\paragraph}[1]{\vspace{.1em}\noindent\textbf{#1}}
\usepackage[font=small]{caption}

\usepackage{url}            
\usepackage{booktabs}       
\usepackage{amsfonts}       
\usepackage{nicefrac}       
\usepackage{microtype}      
\usepackage{lipsum}		
\usepackage{graphicx}
\usepackage{natbib}
\usepackage{doi}
\usepackage{caption}  
\usepackage{textcomp,booktabs}
\usepackage{multirow}
\usepackage{amsmath}
\usepackage{enumitem}

\usepackage{booktabs, multirow, tabularx, siunitx}
\usepackage{tabularx} 
\usepackage{svg}

\makeatletter
\newcommand{\labelline}[1]{%
  \begingroup
    \edef\@currentlabel{\number\value{ALG@line}}%
    \label{#1}%
  \endgroup
}
\newcommand{\lineref}[1]{line~\ref{#1}}
\makeatother

\usepackage{algorithm}
\usepackage{algpseudocode}
\algrenewcommand\algorithmicrequire{\textbf{Input:}}
\algrenewcommand\algorithmicensure{\textbf{Output:}}
\algnewcommand{\LineComment}[1]{\State \(\triangleright\) #1}

\title{MoTVLA: A Vision-Language-Action Model with Unified Fast-Slow Reasoning}

\author{
Wenhui~Huang$^{*1}$\\
\And
Changhe~Chen\thanks{Equal Contribution. Email: \texttt{oscarhuang@seas.harvard.edu}}$^{\ \ 2}$\\
\And
Han~Qi$^{1}$ 
\And 
Chen~Lv$^{3}$
\And
Yilun~Du$^{1}$
\And
Heng~Yang$^{1}$
\And 
$^1$Harvard University \quad  $^2$University of Michigan \quad $^3$Nanyang Technological University
}

\hypersetup{
pdftitle={A template for the arxiv style},
pdfsubject={q-bio.NC, q-bio.QM},
pdfauthor={David S.~Hippocampus, Elias D.~Striatum},
pdfkeywords={First keyword, Second keyword, More},
}

\begin{document}
\maketitle

\vspace{-10mm}

\begin{center}
    \url{https://motvla.github.io/MoTVLA-website/}
\end{center}

\vspace{3mm}

\begin{abstract}
Integrating visual-language instructions into visuomotor policies is gaining momentum in robot learning for enhancing open-world generalization. Despite promising advances, existing approaches face two challenges: limited language steerability when no generated reasoning is used as a condition, or significant inference latency when reasoning is incorporated.
In this work, we introduce MoTVLA, a mixture-of-transformers (MoT)–based vision–language–action (VLA) model that integrates fast–slow unified reasoning with behavior policy learning. MoTVLA preserves the general intelligence of pre-trained VLMs (serving as the generalist) for tasks such as perception, scene understanding, and semantic planning, while incorporating a domain expert, a second transformer that shares knowledge with the pretrained VLM, to generate domain-specific fast reasoning (e.g., robot motion decomposition), thereby improving policy execution efficiency. By conditioning the action expert on decomposed motion instructions, MoTVLA can learn diverse behaviors and substantially improve language steerability. Extensive evaluations across natural language processing benchmarks, robotic simulation environments, and real-world experiments confirm the superiority of MoTVLA in both fast-slow reasoning and manipulation task performance. 

\end{abstract}


\section{Introduction}
Vision-Language-Action (VLA) models, as natural extensions of Vision-Language Models (VLMs), have recently attracted growing interest in robot learning by generating action trajectories through next-token prediction \citep{openvla, ecot}. While promising progress has been made in tasks such as manipulation \citep{cot-vla, rt2} and mobile navigation \citep{navila}, this paradigm faces inherent limitations. Compared to natural language processing (NLP), the scale of robotic datasets is much smaller, and fine-tuning large VLAs on such limited data often degrades the general intelligence acquired during pre-training, reducing adaptability and generalization. Moreover, representing continuous actions with discretized tokens can compromise precision and robustness \citep{pi0.5_ki}.

\begin{figure*}[t]
  \centering
  \includegraphics[width=\textwidth]{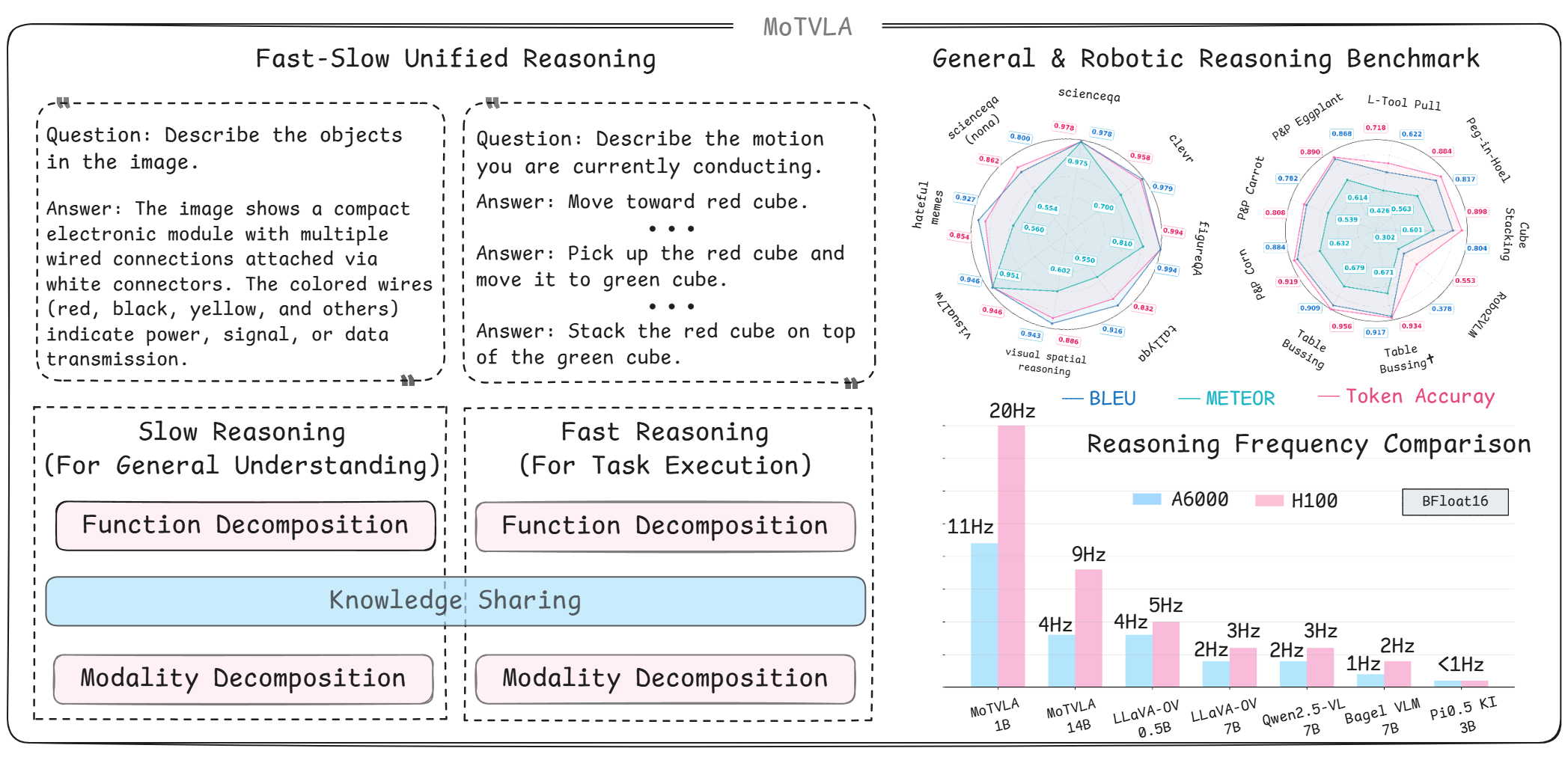}
  \vspace{-10pt}
  \caption{\textbf{Unified fast–slow reasoning in MoTVLA and its reasoning performance.} MoTVLA unifies fast–slow reasoning through a Mixture-of-Transformers architecture, in which the input modalities are first decomposed, followed by knowledge sharing, and finally decomposed again from a functional perspective. As a result, MoTVLA not only learns both general and domain-specific knowledge effectively, but also achieves superior reasoning efficiency.}
  \label{teaser}
  \vspace{-25pt}
\end{figure*}

To address these issues, diffusion policies (DPs) have emerged as a compelling alternative, better suited for modeling continuous action spaces. Leveraging their multimodal nature \citep{ddpm, ddim}, visuomotor DPs capture diverse behaviors via iterative noise–denoise processes \citep{dp}. A recent paradigm integrates VLMs with DPs by conditioning the policies on separately encoded textual and visual features, enabling multitasking through multimodal inputs \citep{rdt}. However, this design suffers from limited language steerability since reasoning is not explicitly generated and conditioned \citep{barreiros2025careful}. Another paradigm employs autoregressive VLMs to generate reasoning in the language domain, with DPs subsequently conditioned on this reasoning for action generation \citep{pi0.5, pi0.5_ki}. While this approach improves behavioral generalization, inference remains constrained by the latency of next-token prediction, which hinders time-critical applications.

To bridge the aforementioned gap, we propose a mixture-of-transformers (MoT)–based \citep{mot} vision-language-action (VLA) model, termed MoTVLA (\textit{pronounced ``MotiLA''}), which integrates fast–slow unified reasoning with policy learning through diffusion policies (DPs). Unlike existing approaches, MoTVLA unifies fast and slow reasoning within a single architecture by decomposing the modalities at the input and the functionalities at the output, while maintaining a shared global knowledge base in between (Fig.~\ref{teaser}). The slow reasoning process follows the standard next-token prediction paradigm of the VLM (generalist), whereas the fast reasoning process is realized via token-wise prediction in a secondary transformer (domain expert) that shares global self-attention with the generalist. For illustration, we refer to these as the first and second transformers; however, in MoTVLA they are integrated into one architecture through a shared global attention mechanism. This design enables MoTVLA to retain the general intelligence of pre-trained VLMs for tasks such as perception, scene understanding, and semantic planning, while efficiently acquiring domain-specific fast reasoning, such as robot motion decomposition, by leveraging globally shared common knowledge. Finally, the action expert, implemented as a diffusion transformer (DiT), is conditioned on the reasoning signals together with visual and physical states to generate language-steered action trajectories, thereby closing the loop from high-level reasoning to low-level control. During inference, MoTVLA operates in two interactive modes: a dialogue mode that allows MoTVLA to answer human queries through slow reasoning by the generalist, and an action mode in which MoTVLA performs specific tasks instructed through fast reasoning and action diffusion by the reasoning backbone and action expert. This dual-mode design ensures that MoTVLA’s responses remain aligned with the language prompt, for example, answering general questions verbally without executing unintended actions, and performing decomposed motion tasks efficiently when receiving task-related prompts.

We conduct comprehensive evaluations of MoTVLA across natural language processing (NLP) benchmarks, robotic simulation environments (ManiSkill) \citep{maniskill}, and real-world experiments. These validations cover a wide spectrum of tasks, ranging from visual understanding on image-based vision–question–answering (VQA) for text and mathematics to robotic manipulation tasks such as stacking, tool usage, insertion, pick-and-place, and table bussing. The results consistently confirm the superiority of MoTVLA over state-of-the-art (SOTA) baselines in both reasoning and manipulation tasks. Finally, the ablation study confirms the significance of the proposed architecture for both reasoning and policy learning.

We summarize our main contributions as follows: 
\begin{enumerate}[label=(\roman*), nosep]
    \item We unify fast and slow reasoning within a single model based on the MoT architecture, enabling the preservation of general intelligence while efficiently learning domain-specific knowledge that benefits from it.
    \item We condition policy learning on decomposed motion generated by fast reasoning, thereby facilitating faster task execution while maintaining interpretability of policy behaviors within a language context. 
    \item MoTVLA achieves superior performance in inference latency, reasoning, and manipulation tasks, providing a novel insight of integrating reasoning into downstream behavior policy.
\end{enumerate}


\paragraph{Outline.} After a review of related work on VLA and DP in Section~\ref{sec:related_work}, we detail the MoTVLA model and training recipe in Section~\ref{sec:model_recipe}. We present experimental results in Section~\ref{sec:exp} and conclude in Section~\ref{sec:conclusion}. Additional details, including inference latency comparisons, dataset descriptions, and further qualitative results, are provided in the Appendix~\ref{sec:appendix} and Supplementary Material.

\vspace{-5pt}
\section{Related Work}\label{sec:related_work}
\vspace{-3pt}

\paragraph{Vision-Language-Action Models.} Vision-language-action (VLA) models, as an important branch of robotic foundation models, have emerged as one of the most prominent approaches for learning multitask policies in robot learning. Owing to their billion-scale parameterization, VLAs are capable of accommodating large-scale robotic datasets \citep{droid, xemb, bridge, robobrain, robo2vlm, llara}. The underlying rationale of this paradigm is to learn behavior policies through next-token prediction, analogous to language modeling, thereby transferring general intelligence into domain-specific knowledge for robotic tasks. Representative examples include RT-2 \citep{rt2} and OpenVLA \citep{openvla}, which pioneered the learning of visuomotor control policies by leveraging vision-language models (VLMs) and modeling continuous actions through discretized action tokens. Recognizing that purely end-to-end training can impair reasoning capabilities, subsequent work such as ECoT \citep{ecot}, Gemini Robotics \citep{geminirobotics}, and CoT-VLA \citep{cot-vla} proposed to jointly learn textual and visual understanding alongside visuomotor policy learning. Despite their demonstrated success across diverse robotic tasks, VLAs face challenges that hinder their practical applications. In particular, control accuracy is often compromised by the information loss incurred when continuous actions are represented with discrete tokens \citep{pi0.5_ki}.

\paragraph{Diffusion Policy.} DPs \citep{dp, rdp, qi2024control} have been widely adopted for robot policy learning by leveraging the strong generative capabilities of diffusion models \citep{ddim, ddpm} in visual generation. The central idea of DP is to model the multimodality of robot behaviors through the noise–denoise process, which is naturally suited for continuous action spaces. Recently, an emerging research direction has focused on advancing diffusion-based VLAs \citep{rdt, tinyvla, pi0.5, pi0.5_ki, graspvla, villa-x, groot, lbm} by integrating VLMs with DP, thereby combining the strengths of both paradigms. For example, RDT-1B \citep{rdt} tokenizes textual and visual inputs, encodes them, and conditions the action diffusion process on this information, resulting in a multitask DP. However, as highlighted by LBM \citep{lbm}, such lightweight integration faces limitations in language steerability because it only encodes input information, which inherently lacks reasoning content. In contrast, the $\pi0.5$ family \citep{pi0.5, pi0.5_ki} first generates textual output based on input images and prompts, and then conditions the DP on this reasoning through flow matching, thereby enabling instruction-following action policies. While these approaches achieve impressive generalization in real-world settings, their reliance on next-token prediction for reasoning introduces significant inference latency, which in turn limits task execution efficiency.

\begin{figure*}[t]
  \centering
  \includegraphics[width=\textwidth]{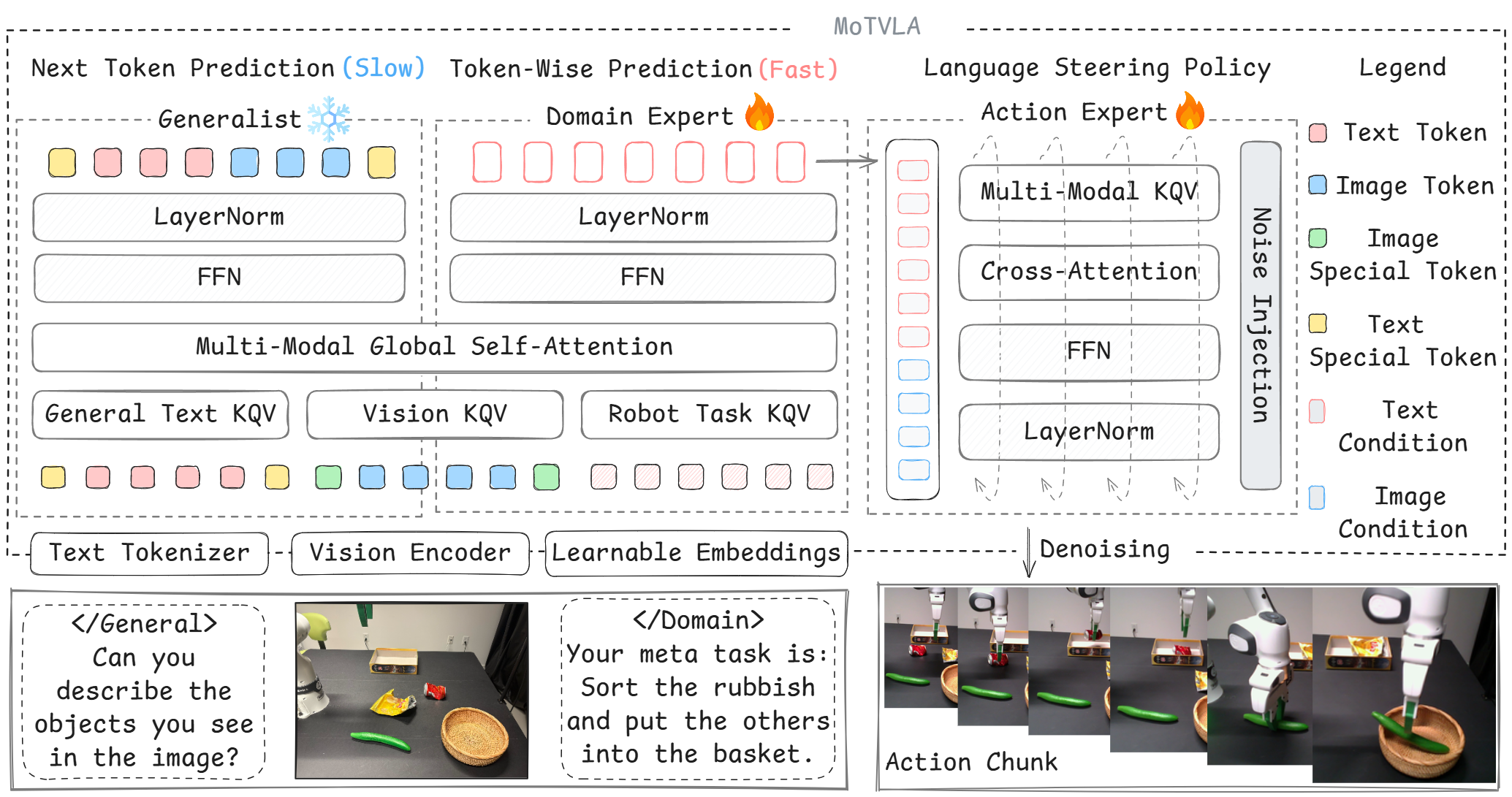}
  \caption{\textbf{The universal framework of MoTVLA.} MoTVLA adopts a Mixture-of-Transformers architecture comprising a generalist, a domain expert, and an action expert. Its reasoning backbone follows a decomposition–composition–decomposition pipeline: multimodal inputs are first processed independently, then integrated through a unified global self-attention mechanism, and finally decoupled at the output to perform slow and fast reasoning via the generalist and domain expert, respectively. The fast reasoning module decomposes robotic motions, and the resulting motion representation, together with visual and physical states, condition the action expert. This design ensures that the learned policy aligns with motion instructions and enhances language steerability, even under ambiguous prompts.}
  \label{arc}
  \vspace{-15pt}
\end{figure*}

\vspace{-5pt}
\section{The MoTVLA Model, Training Recipe, and Inference Pipeline}\label{sec:model_recipe}
\vspace{-3pt}

\subsection{Model Architecture}

The overall architecture of MoTVLA is illustrated in Fig.~\ref{arc}. MoTVLA adopts a Mixture-of-Transformers (MoT) design and consists of three key components: a generalist, a domain expert, and an action expert. The generalist is dedicated to visual–textual multimodal understanding, the domain expert focuses on fast reasoning for robotic tasks, and the action expert is responsible for multitask policy learning.

\paragraph{Input Space Design.} The input modalities of MoTVLA consist of three domains: (1) language, which provides either general or domain-specific prompts, (2) RGB images, and (3) a set of learnable queries conditioned for fast reasoning generation. To process these inputs, MoTVLA employs a text tokenizer and a visual encoder that jointly support both fast and slow reasoning, along with learnable embeddings specifically designed for fast reasoning. Following BAGEL \citep{bagel}, we adopt a Vision Transformer (ViT) as the visual encoder, initialized from SigLIP2-so400m/14 \citep{tschannen2025siglip} with a fixed input resolution of 384. For the text tokenizer, we directly use the one from the pre-trained Qwen2.5 LLM \citep{hui2024qwen2}.

\paragraph{Reasoning Backbone Design: Decomposition-Composition-Decomposition.} To realize fast–slow unified reasoning, we follow the MoT design principle \citep{mot}. Specifically, we adopt Qwen2.5 LLM 7B \citep{hui2024qwen2} as the generalist backbone and mirror the same architecture for the domain expert, which includes RMSNorm \citep{rmsnorm}, RoPE \citep{rope} for positional embedding, and an additional QK-Norm module. Task information is decomposed into the three modalities introduced above and tokenized separately to produce multimodal tokens and their corresponding QKVs. These QKVs are then aggregated into a unified set for joint global attention, with modality-specific masks regulating interaction and conditioning, formulated as:
\begin{equation}
\begin{aligned}
     \mathrm{GA}(x, \{\theta^{m}_{\text{LA}}\}) &= \left( \mathrm{softmax} \left(\frac{QK^T}{\sqrt{d_k}} \right) V \right)W^{m_i}_{QKV} \\
    \text{att} &= \mathrm{GA}(x, \{\theta^{m}_{\text{LA}}\}_{m \in \{\text{text, image, queries}\}}) \\
    h &= x  + \mathrm{LayerNorm}^{m_i}_{\text{LA}}(\text{att}_i) \ ,
\end{aligned}
\end{equation}
where GA denotes the global attention operation within the reasoning backbone, that is, between the generalist and the domain expert, and LA refers to the local attention operation computed independently within each expert. Additionally, $x = {x_1, x_2, ..., x_n}$ represents the input token sequence, $m$ denotes the modality, $\theta$ indicates the trainable parameters, and $W^{m_i}$ corresponds to the modality-specific projection weights. In this formulation, QKVs of later modalities are allowed to attend to those of earlier ones. Within each modality, we apply two types of local attention for text and maintain bidirectional attention for vision. The global attention is then decomposed by modality indices, enabling distinct functions such as general (slow) and domain-specific (fast) reasoning. Following this decomposition–composition–decomposition paradigm, MoTVLA preserves the general intelligence inherited from pre-training by decoupling the associated parameters, while also facilitating fast reasoning through effective knowledge sharing from the generalist to the domain expert. The importance of this design is validated by the ablation study in Section \ref{ablation-study}. Notably, the current design requires the generalist and domain expert to share the same model size, resulting in the MoTVLA reasoning backbone containing twice the parameters of the generalist alone. All primary training and evaluation in this work are conducted with MoTVLA-14B, while the 1B variant is used only to illustrate potential inference speed gains when scaling down. Limitations are discussed in Section \ref{sec:conclusion}, and inference frequency comparisons are provided in Appendix \ref{latency}.

\paragraph{Reasoning Output Design.} The reasoning output of MoTVLA is unified in the textual space but decoupled into two functionalities: slow and fast reasoning. Slow reasoning follows the standard next-token prediction paradigm with causal attention, leveraging the strengths of autoregressive LLMs but incurring high latency. Owing to large-scale pre-training on internet-scale datasets, the generalist demonstrates strong generalization across tasks such as perception, scene understanding, and semantic planning. In contrast, fast reasoning adopts a token-wise prediction paradigm with bidirectional attention, enabling substantially faster text generation. This design allows MoTVLA to pass hidden state inferences from the domain expert directly to the action expert without multiple forward passes. The key insight is that hidden states from a single forward process in token-wise prediction already encode the information required for reasoning generation, whereas those from next-token prediction reflect only the input information. Although token-wise prediction inevitably sacrifices some reasoning accuracy, it is sufficient for producing simple outputs such as decomposed manipulation motions in this work.

\paragraph{Action Expert Design.} In our setting, to better accommodate the varying token lengths of hidden states and their associated masks, we adopt the Diffusion Transformer (DiT) as the action expert of MoTVLA, instead of a U-Net. Policy learning is performed within the framework of action diffusion \citep{dp}, which captures the multimodal nature of robot behaviors. The state space of the action expert consists of four components: (1) visual observations $I_{t-H_I:t}$ with time horizon $H_I$, (2) the motion decomposition representation $h_{t}^{DE}$ generated by the domain expert through fast reasoning, (3) the robot configuration $q_{t-H_I:t}$ (e.g., joint angles and gripper status), and (4) noisy action trajectories ${A}_{t:t+H_{A}}$, where $H_{A}$ denotes the action horizon. The diffusion policy learned by the action expert can be formulated as:
\begin{equation} 
    \hspace{-2mm} \pi_{\theta_{\text{AE}}} (A_{t:t+H_A}, h_{t}^{DE}|I_{t-H_I:t}, \ell, q_{t-H_I:t})\! =\! \pi_{\theta_{\text{AE}}}(A_{t:t+H_A}|I_{t-H_I:t}, h_{t}^{DE}, q_{t-H_I:t}) \pi_{\theta_{\text{RE}}} (h_{t}^{DE}|{I}_t, \ell),
\end{equation}
where $I_t$ represents the image state at timestep $t$, $\ell$ denotes the input prompts, and $H_{\mathcal{I}}$ indicates the time horizon of visual observations. The parameters $\theta_{\text{AE}}$, $\theta_{\text{DE}}$, and $\theta_{\text{RE}}$ correspond to the trainable weights of the action expert, domain expert, and the reasoning backbone (comprising both the generalist and domain expert), respectively.

\begin{figure*}[t]
  \centering
  \includegraphics[width=1.0\textwidth]{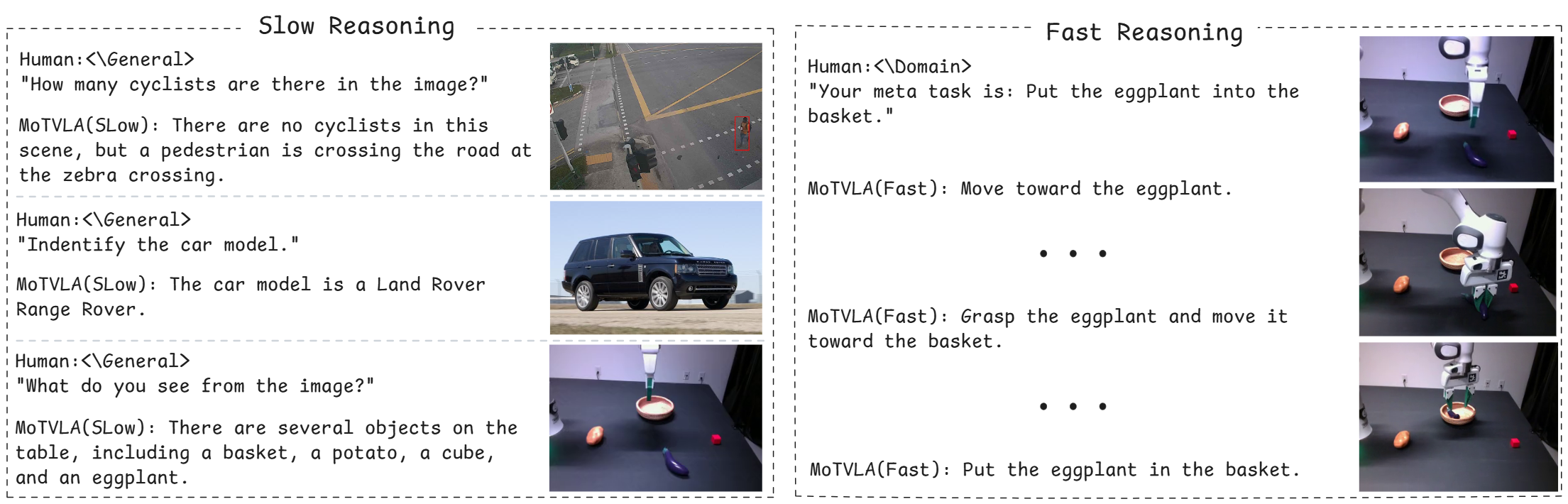}
  \caption{\textbf{A concrete example of fast and slow reasoning during inference.} When asked a general question, the generalist performs slow reasoning and provides a detailed verbal response. When a task-execution prompt is issued, the domain expert performs fast reasoning to generate decomposed intermediate motions for action execution.}
  \label{pic:fast-slow}
  \vspace{-10pt}
\end{figure*}

\subsection{Training Recipe}

Following the abovementioned architecture, MoTVLA has three individual parts, the generalist, domain expert, and action expert, to train with. To leverage strong power of the general intelligence, in this work we adopt the pre-trained VLM from Bagel \citep{bagel}, which achieves SOTA performance on visual understanding benchmarks, as the initialization of the generalist. Therefore, we dedicate our effort to train the rest of two stages in this work. 

\paragraph{Domain Expert Supervised Fine-Tuning.}\label{de-sft}
In the domain expert SFT stage, we construct a high-quality manipulation motion VQA dataset by combining both simulated data and real-world demonstrations from human operators. For data construction, we adopt a question template of the form “Your meta task is: …” and fill the subsequent part with a specific task description, e.g., “Sort the rubbish into the box and move the other into the basket.” For the corresponding answer, we employ the generalist to generate decomposed motions in four steps, which are then used as the training labels for the action expert. To further improve generalization capability, we jointly train the domain expert with two additional open-source datasets: LLaVA-OV \citep{llavaov}, which contributes to language generalization, and Robo2VLM \citep{robo2vlm}, which enhances robotic knowledge. We manually filter and curate long-answer samples from LLaVA-OV to ensure their suitability for token-wise prediction learning, while converting the selection-style annotations in Robo2VLM into reasoning-based labels. In total, our action expert training dataset consists of 1.27M QA pairs, including 154K samples from simulation and 125K from real-world demonstrations collected in-house, 678K from Robo2VLM, and 318K from LLaVA-OV. Further details are provided in Appendix \ref{sft-dataset}. The learning objective is to minimize the negative log-likelihood of target tokens:
\begin{equation}
    \mathcal{L}(\theta_{\text{DE}}) = \mathbb{E}_{(x,y) \sim D}[-\log p_{\theta}(y_{1:n}|x_{1:n})] \,
\end{equation}

\noindent where $D$ denotes dataset and $y$ represents sequence of reasoning tokens in this stage.

\paragraph{Action Expert Diffusion Policy.} 
We collect $300$ demonstrations in the ManiSkill \citep{maniskill} simulator for each of three tasks (Cube Stacking, Peg-in-Hole, and L-tool Pull), with additional distractors introduced into the scenes. In the real world, using the SpaceMouse, we collect $50$ demonstrations for pick-and-place with various vegetables and $200$ demonstrations for table bussing. Notably, this dataset is substantially smaller than that used for training the domain expert, as publicly available datasets with decomposed motion annotations are hardly available, e.g. $\pi$0.5 \citep{pi0.5}, and the workload of collecting and annotating such data in-house is prohibitively large. Conditioning on the visual observations $I_{t-H_I:t}$ over $H_I=5$ observation horizons, robot configuration $q_{t-H_I:t}$, and decomposed motion signal $h_{t}^{DE}$, we implement a visuomotor policy using a conditional denoising diffusion model \citep{dp}. During testing, the policy denoises Gaussian noise into action trajectories ${A}_{t:t+H_A}^0$ consisting of $H_A$ steps starting at time $t$. Specifically, beginning from Gaussian noise ${A}_{t:t+H_A}^K$, the denoising network $\epsilon_\theta$ iteratively refines the actions over $K$ denoising steps until producing the noise-free action ${A}_{t:t+H_A}^0$, computed as:
\begin{equation}
    {A}_{t:t+H_A}^{k-1} = \bar{\alpha}_k \left( {A}_{t:t+H_A}^k - \epsilon_\theta \!\left( {A}_{t:t+H_A}^k, k, I_{t-H_I:t}, q_{t-H_I:t}, h_{t}^{DE} \right) \right) + \bar{\beta}_k \, \epsilon^k, 
\end{equation}
where $\epsilon^k \sim \mathcal{N}(0, \mathbf{I})$ and $\bar{\alpha}_k, \bar{\beta}_k$ are noise scheduler coefficients.

To train the denoising network $\epsilon_\theta$, we corrupt the ground-truth action ${A}_{t:t+H{a}}^0$ with noise $\epsilon^k$ at the $k$-th step, and optimize the network to predict the injected noise \citep{dp}, expressed as:
\begin{equation}
    \label{equation:training_objective}
    \mathcal{L}(\theta_{\text{AE}}) = \mathrm{MSE}\!\left( \epsilon^k, \, \epsilon_\theta \!\left( \bar{\alpha}_k {A}_{t:t+H_A}^0 + \bar{\beta}_k \epsilon^k, \, k, I_{t-H_I:t}, q_{t-H_I:t}, h_{t}^{DE} \right) \right).
\end{equation}

\begin{figure}[t]
\centering
\begin{minipage}{\linewidth}
\begin{algorithm}[H]
  \caption{MoTVLA Inference: Generalist $\rightarrow$ Domain Expert $\rightarrow$ Action Expert}
  \label{alg:motvla_infer}
  \begin{algorithmic}[1]
    \Require Models: \(\,GE\) (Generalist/General Expert), \(DE\) (Domain Expert), \(AE\) (Action Expert)
    \Require $I_0$ \(\triangleright\) initial image, $\ell$ \(\triangleright\) general prompt, $\ell_{T}$ \(\triangleright\) task prompt, $H_I$ \(\triangleright\) obs-horizons, \,$H_A$ \(\triangleright\)action-horizons
    \Ensure Closed-loop action sequence

    \Procedure{MoTVLA\_Inference}{$I_0,\,\ell$}
      \State $\textsc{RecentObservations} \gets [\,]$
      \State $t \gets 0$;\;
    \If{$\ell$ is not None}
      \State $Diaglogue \gets GE.\textsc{SlowReason}(I_0,\,\ell)$  \Comment{multi-turn dialogue} \labelline{taskprompt} 
    \ElsIf{$\ell_{T}$ is not None}
      \State $\textsc{RecentObservations.update}(I_0, q_0)$
      \While{$t < \textsc{max\_steps}$}
        \State $h^{DE}_t \gets DE.\textsc{FastReason}(I_t,\, \ell_{T})$ \Comment{fast reasoning, step-wise motion decomposition;} \labelline{motionprompt}
        \State $I_{t-H_I:t}, q_{t-H_I:t}\gets \textsc{RecentObservations}$  \labelline{sampling}
        \State ${A}_{t:t+H_A}^0 \gets AE.\textsc{Denoise}(I_{t-H_I:t},\, q_{t-H_I:t},\, h^{DE}_t;\, \text{horizon}=H_A)$ \labelline{sample}
        \State \textsc{RobotController}.\textsc{Step}(${A}_{t:t+H_A}^0$) \labelline{execute}
        \State $\textsc{RecentObservations.update}(I_t, q_t)$
        \State $t \gets t + 1$
      \EndWhile
      \State \textsc{Terminate}()
    \EndIf
    \EndProcedure
  \end{algorithmic}
\end{algorithm}
\end{minipage}
\vspace{-15pt}
\end{figure}

\subsection{Inference Pipeline}
Integrating all components into a unified framework, MoTVLA operates in two interaction modes during inference: (i) the operator may engage in multi-turn dialogue with MoTVLA, posing questions or logical queries such as object descriptions or potential risk analysis, and (ii) the operator may instruct MoTVLA to execute manipulation tasks through task-specific prompts. MoTVLA employs slow reasoning when a detailed verbal response is required, and fast reasoning for action execution when task-related prompts are received. This dual-mode design ensures proper alignment between instruction and response.  For example, when asked ``What do you see in the image?'' MoTVLA provides a detailed verbal response through slow reasoning without performing unintended actions. In contrast, when instructed to perform a specific task, MoTVLA activates fast reasoning to generate decomposed intermediate motions for subsequent action diffusion, thereby enhancing both interpretability and language steerability. Figure~\ref{pic:fast-slow} demonstrates a concrete instance of slow and fast reasoning during the inference time.


During manipulation rollouts, the system maintains sliding windows of the most recent \(H_I\) images and robot states to provide contextual awareness (\lineref{sampling}). At each step, the Domain Expert (DE) performs fast reasoning with step-wise motion decomposition on the current image \(I_t\), conditioned on the \(task\_prompt\), and produces a decomposed motion representation \(h_{t}^{DE}\) (\lineref{motionprompt}). The Action Expert (AE) then samples an action chunk \(A_{H_A}^{0}\), conditioned on the image window \(I_{t-H_I:t}\), state window \(q_{t-H_I:t}\), and the decomposed motion representation \(h_{t}^{DE}\) (\lineref{sample}). The resulting action chunk \(A_{H_A}^{0}\) is subsequently sent to the \textsc{RobotController} for execution (\lineref{execute}).

\vspace{-5pt}
\section{Experiments}\label{sec:exp}
\vspace{-3pt}

MoTVLA is a multitasking model capable of performing general slow reasoning, multimodal robot-specific fast reasoning, and action planning for manipulation tasks. We conduct extensive evaluations in both simulation and real-world experiments, covering semantic fast-slow reasoning as well as embodied action execution. 

\subsection{Metrics and Baselines.}
In this work, we employ standard NLP metrics, including BLEU \citep{bleu}, METEOR \citep{meteor}, CIDEr \citep{cider}, and token accuracy, to evaluate reasoning performance. For manipulation tasks, we report the average success rate using random seeds that were not observed during data collection or training.

We compare MoTVLA against several well-known and SOTA baselines, including transformer-based DP \citep{dp}, GR-MG \citep{grmg}, $\pi0$ \citep{pi0}, and the recently released $\pi0.5$ with knowledge insulation \citep{pi0.5_ki}. For a fair comparison and to align the baselines’ output space with our hardware setting, we fine-tuned them on our own dataset, which contains 1,050 trajectories, on top of their respective pre-trained models.

\begin{table*}[tp]
\caption{Fast reasoning evaluation on both robotics and LLaVA-OV VQA tasks. ($*$ refers to revision, $^\dagger$ denotes the same task evaluated under an alternative instruction prompt.)}
\label{fastreasoning}
\centering
\small
\resizebox{\textwidth}{!}{
\begin{tabular}{c | c | c c c c }
\toprule
\multirow{2}{*}{\textbf{Task}} & \multirow{2}{*}{\textbf{Dataset}} &
\multicolumn{4}{c}{\textbf{Metrics}} \\
\cmidrule(lr){3-6} 
 &  &
\textbf{BLEU (0-1)$\uparrow$} & \textbf{CIDEr (0-10) $\uparrow$} & \textbf{METEOR (0-1)$\uparrow$} & \textbf{Token Accuracy (\%)$\uparrow$} \\
\midrule
\multirow{9}{*}{Robotics} & Cube Stacking&
0.8041 & 8.0574 & 0.6005 & 89.82 \\
 & Peg-in-Hole &
0.8174 & 7.6805 & 0.5633 & 88.41 \\
 & L-tool Pull &
0.6221 & 6.6255 & 0.4263 & 71.76  \\
 & Pick \& Place: Eggplant & 
 0.8680 & 8.3945 & 0.6136 & 88.99 \\
 & Pick \& Place: Carrot &
0.7816 & 7.1222 & 0.5395 & 80.83 \\
 & Pick \& Place: Corn &
 0.8836 & 8.7838 & 0.6320 & 91.86 \\
 & Table Bussing &
0.9086 & 8.9802 & 0.6794 & 95.59 \\
 & Table Bussing$\dagger$ &
0.9168 & 9.0454 & 0.6712 & 93.41 \\
 & Robo2VLM$^*$ &
0.3782 & 2.3601 & 0.3279 & 57.05 \\
\midrule
\multirow{8}{*}{LLaVA-OV VQA} &
 FigureQA~\citep{figureqa} & 0.9940 & 2.4850 & 0.8105 & 99.40 \\
 & CLEVR~\citep{clevr} & 0.9790 & 2.3950 & 0.7004 & 95.80\\
 & ScienceQA~\citep{scienceqa} & 0.9780 & 2.4450  & 0.9749 & 97.80 \\
 & ScienceQA(nona context)~\citep{llavaov} & 0.8004 & 2.4282 & 0.5544 & 86.16 \\
 & Hateful-memes~\citep{hatefulmemes}  & 0.9270 & 2.1350 & 0.5603 & 85.40 \\
 & Visual-7W~\citep{visual7w} & 0.9459 & 2.3647 & 0.9510 & 94.59 \\
 & Visual Spatial Reasoning~\citep{vsr} & 0.9430 & 2.2150 & 0.6022 & 88.60 \\
 & TallyQA~\citep{tallyqa} & 0.9160 & 2.0850 & 0.5498 & 83.20 \\
\bottomrule
\end{tabular}
}
\end{table*}

\begin{figure*}[t]
  \centering
  \includegraphics[width=0.9\textwidth]{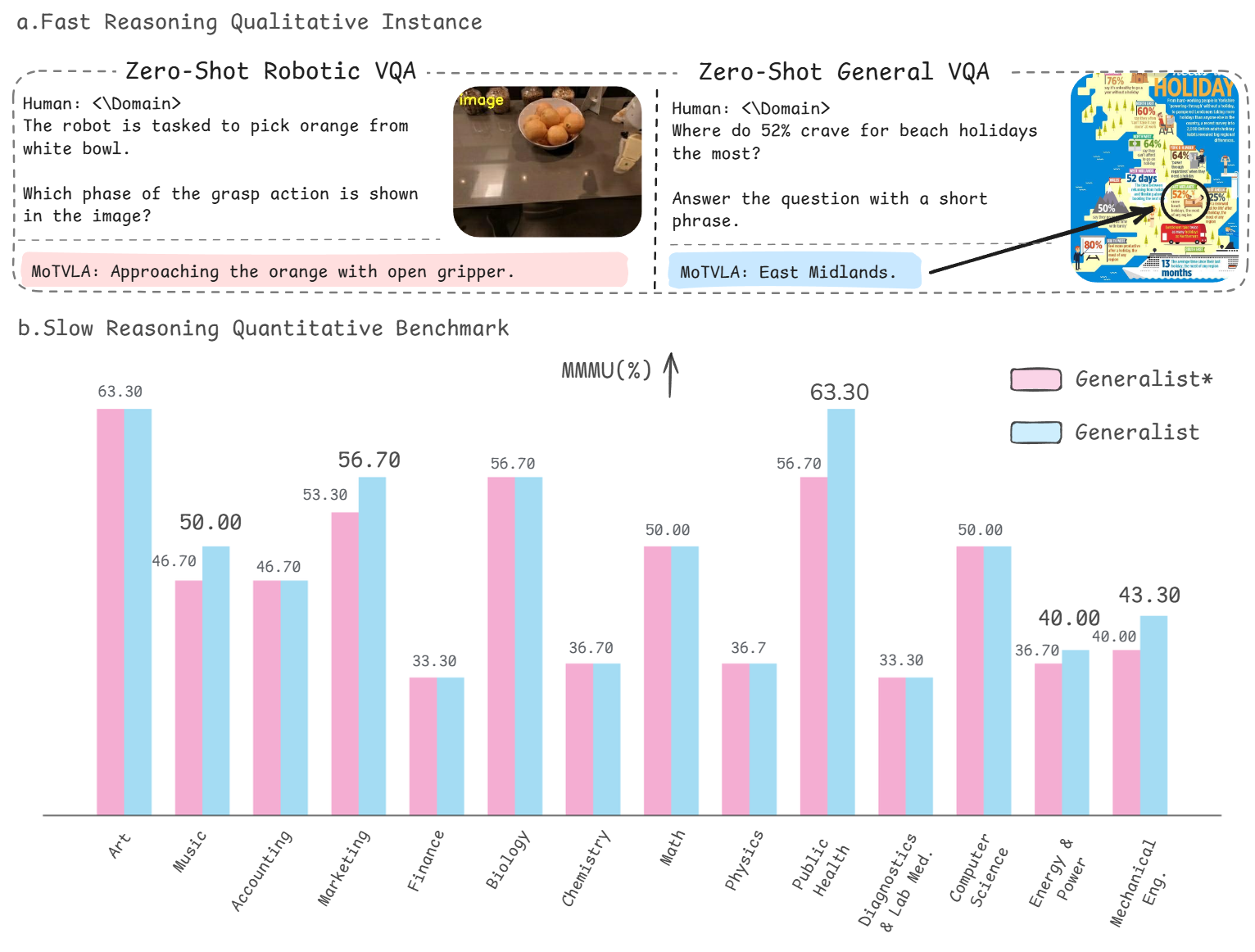}
  \vspace{-5pt}
  \caption{\textbf{Fast reasoning instances and slow reasoning benchmark.} $*$ indicates that the generalist is fine-tuned on the data originally intended for training the domain expert.}
  \label{qualitative-reasoning}
\vspace{-5pt}
\end{figure*}

\subsection{Reasoning Tasks}\label{subsec:reasoning_tasks}

The performance of fast reasoning is particularly critical in our work, since its outputs serve as conditioning signals for the action expert. A completely hallucinated reasoning would misguide the diffusion policy and lead to undesirable behaviors. As discussed in Section \ref{de-sft}, we incorporate additional VQA datasets, namely LLaVA-OV~\citep{llavaov} and Robo2VLM \citep{robo2vlm}, in addition to our own robotic motion reasoning dataset, to enhance generalization capability. Distinct from existing VLA approaches, we not only qualitatively assess the quality of fast reasoning through VQA, but also quantitatively evaluate its performance on both robotic and general datasets.

\paragraph{Analysis.} 
The overall statistical results of fast reasoning are summarized in Table \ref{fastreasoning}. The average BLEU score and token accuracy across robotics and LLaVA-OV VQA tasks highlight its superior precision in capturing both domain expertise and common knowledge. Meanwhile, the CIDEr and METEOR scores on robotics tasks exhibit strong alignment with human judgment in terms of expression diversity and semantic similarity, indicating that the motion decomposition generated through fast reasoning has been effectively learned. The manually curated Robo2VLM reasoning dataset further underscores the difficulty of this evaluation, as we reformatted it from multiple-choice to reasoning-based VQA, encompassing diverse scenes, tasks, views, and reasoning domains (spatial, goal-oriented, and interaction-based). For LLaVA-OV VQA, although the CIDEr score is less prominent and below human-level quality, the relatively high METEOR score demonstrates sufficient generalization, justifying our motivation for joint training with both general and domain-specific datasets.

\looseness=-1
Qualitative evaluations on two randomly selected reasoning tasks (Fig.\ref{qualitative-reasoning}a) further confirm strong zero-shot generalization. Surprisingly, MoTVLA is able to infer unseen objects and decompose motions never encountered during training, and can also extract key information from an information-dense poster. These results collectively demonstrate MoTVLA’s strong domain knowledge and generalization capability to handle diverse scenarios. Detailed reasoning latency comparisons and more qualitative results are provided in Appendix~\ref{latency} and supplementary materials due to limited space.

Furthermore, we conduct an ablation study on slow reasoning for general knowledge evaluation by comparing training versus freezing the generalist, demonstrating the necessity of the MoT architecture in our approach. To this end, we fine-tuned the generalist using the data originally intended for training the domain expert and compared it with the unfine-tuned version. Since the generalist of MoTVLA is initialized from Bagel, we benchmark both versions on the Massive Multi-discipline Multimodal Understanding and Reasoning (MMMU) dataset~\citep{mmmu}, following the reasoning benchmark employed in the original Bagel paper. As shown in Fig.~\ref{qualitative-reasoning}b, after fine-tuning, the performance of slow reasoning degrades across several subjects. For instance, while the tuned generalist maintains its performance in domains such as art, accounting, finance, and other science-related subjects, it exhibits knowledge forgetting in music, marketing, energy and power, and mechanical engineering. This degradation is even more pronounced in public health, where accuracy drops by 6.6\% after fine-tuning. We also observe similar catastrophic forgetting in robotic VQA when providing a meta-task description followed by a general question, for example: “Your meta task is: Put the cucumber into the box. Before executing the task, tell me what is the color of the cucumber.” The fine-tuned generalist fails to respond to the question about color and instead outputs the decomposed motion. These results highlight the superiority of MoTVLA, which preserves slow reasoning capability even after robotic VQA learning, confirming the necessity of the MoT architecture for maintaining fast–slow reasoning performance.

Overall, the demonstrated superiority of fast reasoning performance establishes a solid foundation for MoTVLA to effectively learn robotic knowledge and corresponding behavior policies while maintaining its general intelligence. The policy learning aspect will be further analyzed, both qualitatively and quantitatively, in the following sections.

\begin{figure*}[t]
  \centering
  \includegraphics[width=\textwidth]{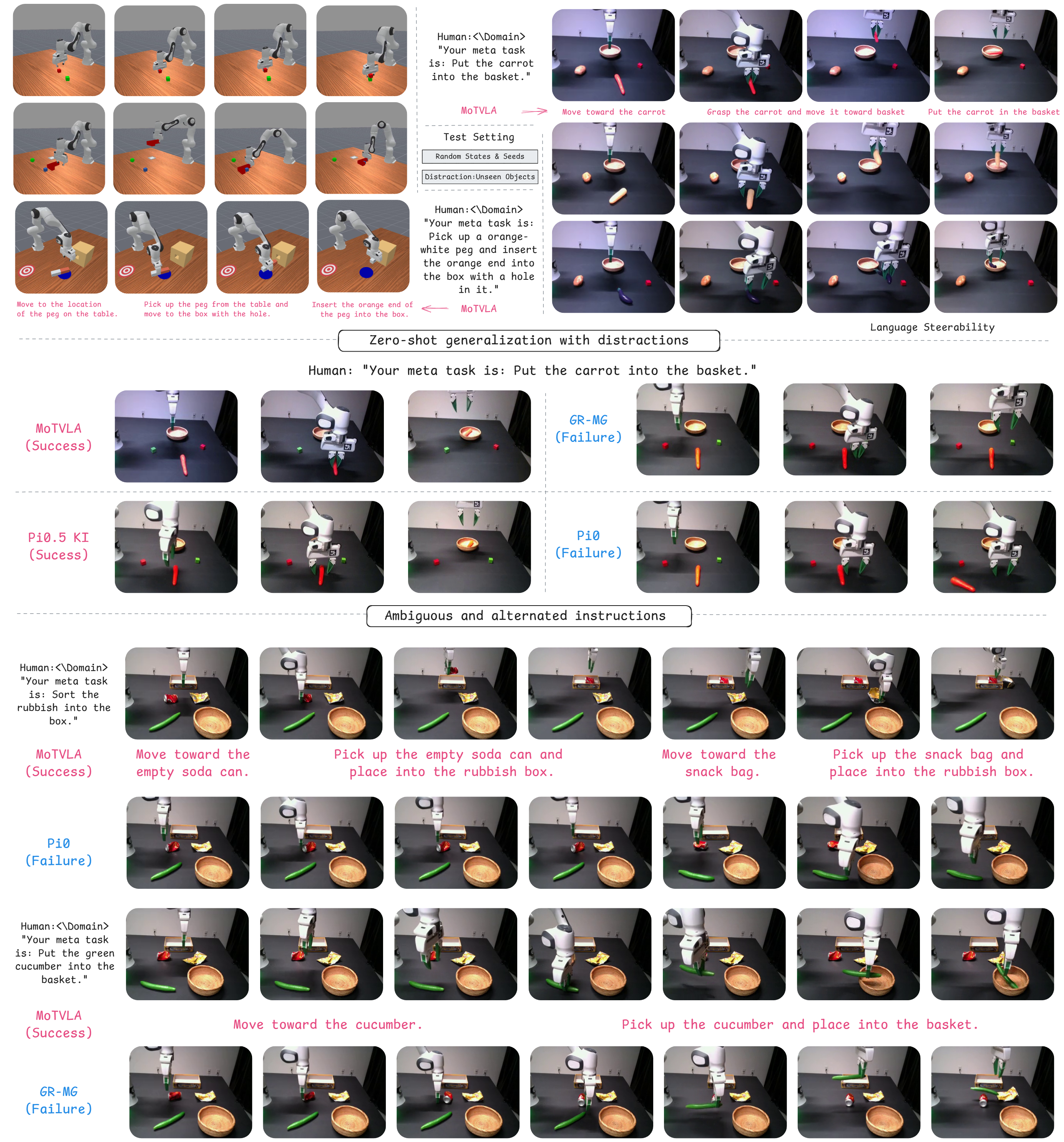}
  \caption{\textbf{Evaluation for manipulation tasks.} MoTVLA is rigorously evaluated in both simulation and real-world experiments. The testing suite encompasses diverse case types and variations, including ambiguous instruction prompts that require strong reasoning capability and language-steered behavioral policies. 
  }
  \label{experiment}
  \vspace{-15pt}
\end{figure*}

\begin{table}[t]
\small
\centering
\caption{Performance comparison across different models for manipulation tasks.}
\resizebox{1.0\textwidth}{!}{
\begin{tabular}{c|ccccc}
\toprule
\textbf{Tasks} & \textbf{MoTVLA} & \textbf{$\pi_{0.5}$ KI} & \textbf{$\pi_0$} & \textbf{GR-MG} & \textbf{DP} \\
\midrule
Cube Stacking                & \textbf{0.79} & 0.30  & 0.02 & 0.14& 0.58 \\
Peg-in-Hole                  & \textbf{0.40} & 0.22  & 0.0 & 0.06 & 0.24 \\
L-Tool Pull                  & \textbf{0.62} & 0.36  & 0.28 & 0.10 & 0.48\\ 
Pick \& Place Eggplant       & \textbf{1.0}  & \textbf{1.0}  & \textbf{1.0}  & 0.50 & 0.25\\
Pick \& Place Corn           & \textbf{1.0}  & \textbf{1.0}  & 0.75 & 0.0 & 0.50  \\
Pick \& Place Carrot         & \textbf{1.0}  & \textbf{1.0}   & \textbf{1.0}    & 0.0 & 0.50  \\
Carrot w/ distractions    & \textbf{1.0} & \textbf{1.0} & 0.75 & 0.0 & 0.75 \\
Eggplant w/ distractions  & 0.75 & \textbf{1.0} & 0.75& 0.0 & 0.50  \\
Corn w/ distractions     & 0.75 & \textbf{1.0} & 0.50 & 0.0 & 0.50  \\
\midrule
Mean $\pm$ Variance           & \textbf{0.81} $\pm$ 0.04 & 0.76 $\pm$ 0.11 & 0.56 $\pm$ 0.11 & 0.09 $\pm$ 0.03 & 0.48 $\pm$ 0.02 \\
\midrule
\textbf{Action Training Recipe} & & & \textbf{Dataset and Scale} & &\\
\midrule
Fine-tuning &  &  & 1050 Collected Trajectories &  & \\
\midrule
\multirow{2}{*}{Pre-training}
& \multirow{2}{*}{None} & 400h + much larger number data + & $\simeq$ 10000h + OXE  & Ego4d ($>$ 3500h)  & \multirow{2}{*}{None} \\
& & OXE \citet{pi0.5} & \citet{pi0} & \citet{grmg}  & \\

\bottomrule
\end{tabular}
\label{p&p}
}
\vspace{-15pt}
\end{table}

\subsection{Robot Manipulation tasks}

We investigate whether MoTVLA can transform language-described motion decompositions into reliable manipulation across both simulation and real-world settings. Our objectives are threefold: (i) to assess whether motion-decomposed condition improves policy learning and robustness in contact-rich tasks, (ii) to examine whether decomposed motion snippets benefit language steerability of the policy behavior, and (iii) to thoroughly benchmark MoTVLA against four strong and SOTA baselines across an evaluation spectrum ranging from simulation to real-world experiments.To this end, we adopt three challenging tasks from the ManiSkill environment \citep{maniskill} and manually increase their difficulty: object relocation with stability (Cube Stacking), tight-tolerance insertion (Peg-in-Hole), and tool-mediated manipulation beyond reach (L-tool Pull). In addition, two types of real-world experiments are employed for training MoTVLA, including three single-stage manipulation tasks with clear instructions and one multi-stage (semi-long-horizon) task with ambiguous instructions. During testing, these tasks are further diversified into additional cases by varying initial states, random seeds, and introducing unseen objects as distractions, as illustrated in Fig.~\ref{experiment}. For example, MoTVLA must distinguish carrots, corn, and eggplants (targets) from potatos (distractions) in a zero-shot manner and guide the policy to complete the task accordingly. Full implementation details are provided in Appendix \ref{annotation}. For fair comparison and to unify the baselines’ output space to our hardware setting, we fine-tuned all baselines on our dataset.

\paragraph{Analysis.} 
The quantitative comparison with baseline methods is reported in Table \ref{p&p}. As shown, MoTVLA consistently outperforms most baselines in both in-domain and zero-shot experiments (with distractions), demonstrating strong robustness and generalization capability. Although the success rate on the challenging \textit{Peg-in-Hole} task is relatively lower than that of other tasks due to its higher precision requirements and tight tolerances, MoTVLA still achieves the best performance among all methods. Furthermore, models with VLM backbones (MoTVLA, $\pi_{0.5}$ KI, $\pi_0$) consistently surpass those without pre-training, underscoring the importance of general intelligence in real-world robotic tasks. The $\pi_{0.5}$ KI model achieves performance comparable but slightly lower overall, while surpassing ours on real-world experiments under zero-shot distractions. This is reasonable, as it was finetuned from a pre-trained checkpoint already trained on large-scale real-world robotic datasets, as indicated in Table \ref{p&p}. However, the remaining baseline approaches such as GR-MG and $\pi_0$ exhibit unstable performance when encountering unseen distractions during zero-shot generalization evaluation. As shown in the middle section of Fig.~\ref{experiment}, the grasping behaviors of the baselines are significantly affected even by minor pixel-level alterations (such as the small red and green cubes), whereas MoTVLA handles such cases robustly. This robustness arises because the reasoning backbone can generalize to unseen scenarios and generate accurate decomposed motion representation, which in turn drive the behavior policy to remain stable under perturbations.


To verify language-steerability of MoTVLA, we design a relevant task, namely table bussing, in which only an ambiguous prompt, “Sort the rubbish into the box,” is provided without specifying which objects constitute rubbish. The qualitative results are presented in the lower part of Fig.~\ref{experiment}. As shown, MoTVLA successfully decomposes the task into several reasonable motions and guides the policy to complete the instruction, whereas $\pi_0$ fails in this task by treating all objects as rubbish and wandering back and forth among them. Moreover, we provide an alternative prompt in the same scene and observe that the policy behavior effectively aligns with the language instruction, thereby demonstrating the superiority of language steerability. Additional qualitative results on manipulation tasks are provided in supplementary materials.


\begin{figure*}[t]
  \centering
  \includegraphics[width=\textwidth]{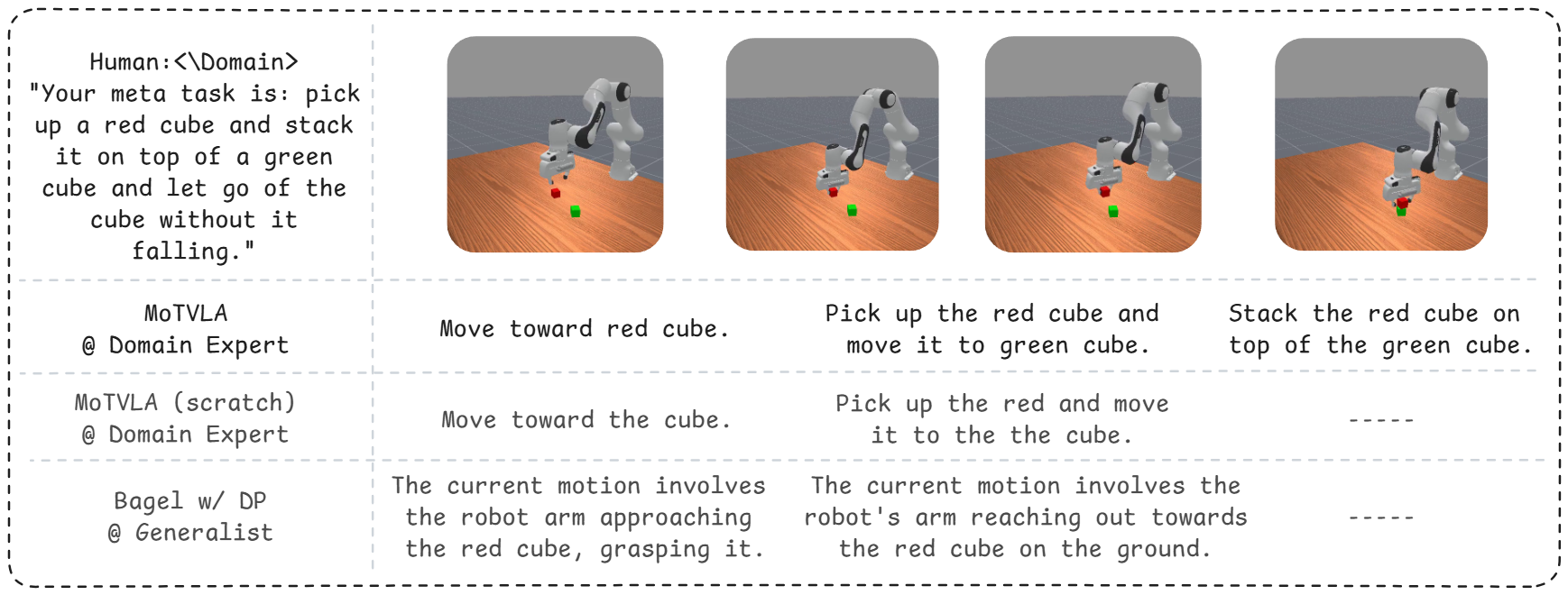}
  \caption{\textbf{A concrete qualitative instance for architecture ablation study.} MoTVLA is able to complete the cube stacking taskm, while MoTVLA (scratch) and Bagel w/ DP cannot due to the improper motion reasoning.
  }
  \label{pic:ablation}
  \vspace{-5pt}
\end{figure*}

\begin{table*}[tp]
\caption{\textbf{Ablation study} of the architectural design. 
P\&P represents pick and place, TA denotes token accuracy, and SR refers to success rate.}
\label{ablation}
\centering
\scriptsize
\begin{tabularx}{\textwidth}{
  c | c | 
  >{\centering\arraybackslash}X 
  >{\centering\arraybackslash}X 
  >{\centering\arraybackslash}X }
\toprule
&  &
\textbf{Bagel w/ DP} & \textbf{MoTVLA (scratch)} & \textbf{MoTVLA (ours)} \\
\midrule
\multirow{3}{*}{Component} & Generalist (Slow Reasoning) &
Pre-trained & Scratch & Pre-trained \\
 & Domain Expert (Fast Reasoning) &
N/A & Fine-tuned & Fine-tuned \\
 & Action Expert (Diffusion Policy) &
Fine-tuned & Fine-tuned & Fine-tuned  \\
\midrule
\multirow{2}{*}{Slow Reasoning} 
 & ScienceQA (TA) & 94.12 & 0.0 & 94.12 \\
 & P\&P Motion Decomposition (TA) & 1.68 & 0.0 & 1.68 \\
\midrule
\multirow{2}{*}{Fast Reasoning} 
 & ScienceQA (TA) & N/A & 17.00 & 90.99 \\
 & P\&P Motion Decomposition (TA) & N/A & 12.08 & 82.41 \\
\midrule
\multirow{1}{*}{Diffusion Policy} 
 & Cube Stacking (SR) & 0.0 & 0.0 & 78.00 \\
\bottomrule
\end{tabularx}
\vspace{-5pt}
\end{table*}

\paragraph{Ablation Study.} \label{ablation-study}
In this ablation study, we evaluate the impact of the pre-trained generalist and the necessity of the domain expert. We compare three variants: \textbf{Bagel \citep{bagel} w/ DP} who has only generalist and action expert pre-trained and fine-tuned respectively, enabling slow reasoning and diffusion policy but absent fast reasoning, \textbf{MoTVLA (scratch)} with randomly initialized the generalist and fine-tuned domain and action expert, thus able to perform both fast-slow reasoning and diffusion policy but lack general intelligence, and \textbf{MoTVLA (ours)} proposed by this work.  The purpose of these baselines is twofold: (i) to investigate the significance of general intelligence by comparing our method with MoTVLA (scratch), and (ii) to illustrate the importance of the domain expert, which enables global attention, by comparing against Bagel w/ DP, where such global attention is absent. 



As shown in Table~\ref{ablation}, MoTVLA (scratch) fails to learn across all tasks under both slow and fast reasoning, confirming that the general intelligence inherited from the pre-trained VLM is essential for effective multimodal reasoning. Meanwhile, Bagel w/ DP performs well on ScienceQA but achieves the lowest token accuracy in the slow reasoning of P\&P motion decomposition, indicating that the domain expert and its global attention mechanism are vital for aligning domain-specific knowledge and stabilizing motion generation. Both baselines achieve zero success on manipulation tasks, as unstable and hallucinated motion decomposition introduces significant variance and disrupts policy learning. Figure~\ref{pic:ablation} presents a concrete rollout example with the corresponding motion decomposition. It is evident that the improper and hallucinated reasoning in both baselines causes them to fail to achieve the task goal until the final motion step. While it is possible to fine-tune the generalist of Bagel together with our domain expert to enhance the performance of P\&P motion decomposition in slow reasoning, this approach would lead to catastrophic forgetting of general intelligence, as discussed in Section~\ref{subsec:reasoning_tasks}. In contrast, our proposed MoTVLA achieves the best performance across all tasks through the synergistic collaboration of the generalist, domain expert, and action expert, demonstrating the effectiveness of the proposed architectural design.

\vspace{-5pt}
\section{Conclusion}\label{sec:conclusion}
\vspace{-5pt}

In this paper, we address the challenging gap between reasoning latency and language steerability by proposing MoTVLA, a Mixture-of-Transformers (MoT) architecture-based robotic foundation model that unifies fast and slow reasoning and enhances the language steerability of behavior policies. MoTVLA acquires domain expertise and action policies through a two-stage curriculum learning scheme while preserving the general intelligence inherited from a pretrained VLM throughout the entire training process. Comprehensive benchmarking across semantic reasoning, simulation, and real-world experiments confirms the feasibility and superiority of the proposed approach. We believe that this integration of high-level reasoning and low-level control policies within a unified VLA framework paves the way for advancing robotic learning toward open-ended, language-instructed tasks and language-steered behaviors in large-scale open-world environments.  

Despite its superior performance on most evaluation tasks, the full potential of MoTVLA has not yet been fully explored. For example, we observe that inference speed can be significantly improved when both the generalist and domain expert are scaled down to 0.5B parameters. However, pre-training a 0.5B model to acquire general intelligence remains highly challenging, as it requires multi-stage training with large-scale VQA datasets \citep{llavaov, hui2024qwen2}. In addition, the relatively limited amount of data available for training the action expert sometimes leads to strong reasoning ability but insufficient execution capability, resulting in failures on long-horizon tasks. Leveraging large-scale open-sourced robotics datasets, annotating motion decompositions, and jointly pre-training the action expert with the reasoning backbone will be important directions for future work.  

\subsubsection*{Acknowledgments}
Han Qi and Heng Yang are partially supported by Office of Naval Research grant N00014-25-1-2322. Yilun Du gratefully acknowledges support from PickleRobotics.

\clearpage

\bibliographystyle{unsrtnat}
\bibliography{references}  

\begin{thebibliography}{47}
\providecommand{\natexlab}[1]{#1}
\providecommand{\url}[1]{\texttt{#1}}
\expandafter\ifx\csname urlstyle\endcsname\relax
  \providecommand{\doi}[1]{doi: #1}\else
  \providecommand{\doi}{doi: \begingroup \urlstyle{rm}\Url}\fi

\bibitem[Kim et~al.(2025)Kim, Pertsch, Karamcheti, Xiao, Balakrishna, Nair, Rafailov, Foster, Sanketi, Vuong, et~al.]{openvla}
Moo~Jin Kim, Karl Pertsch, Siddharth Karamcheti, Ted Xiao, Ashwin Balakrishna, Suraj Nair, Rafael Rafailov, Ethan~P Foster, Pannag~R Sanketi, Quan Vuong, et~al.
\newblock Openvla: An open-source vision-language-action model.
\newblock In \emph{Conference on Robot Learning}, pages 2679--2713. PMLR, 2025.

\bibitem[Zawalski et~al.(2025)Zawalski, Chen, Pertsch, Mees, Finn, and Levine]{ecot}
Micha{\l} Zawalski, William Chen, Karl Pertsch, Oier Mees, Chelsea Finn, and Sergey Levine.
\newblock Robotic control via embodied chain-of-thought reasoning.
\newblock In \emph{Conference on Robot Learning}, pages 3157--3181. PMLR, 2025.

\bibitem[Zhao et~al.(2025)Zhao, Lu, Kim, Fu, Zhang, Wu, Li, Ma, Han, Finn, Handa, Lin, Wetzstein, Liu, and Xiang]{cot-vla}
Qingqing Zhao, Yao Lu, Moo~Jin Kim, Zipeng Fu, Zhuoyang Zhang, Yecheng Wu, Zhaoshuo Li, Qianli Ma, Song Han, Chelsea Finn, Ankur Handa, Tsung-Yi Lin, Gordon Wetzstein, Ming-Yu Liu, and Donglai Xiang.
\newblock Cot-vla: Visual chain-of-thought reasoning for vision-language-action models.
\newblock In \emph{Proceedings of the IEEE/CVF Conference on Computer Vision and Pattern Recognition (CVPR)}, pages 1702--1713, June 2025.

\bibitem[Zitkovich et~al.(2023)Zitkovich, Yu, Xu, Xu, Xiao, Xia, Wu, Wohlhart, Welker, Wahid, Vuong, Vanhoucke, Tran, Soricut, Singh, Singh, Sermanet, Sanketi, Salazar, Ryoo, Reymann, Rao, Pertsch, Mordatch, Michalewski, Lu, Levine, Lee, Lee, Leal, Kuang, Kalashnikov, Julian, Joshi, Irpan, Ichter, Hsu, Herzog, Hausman, Gopalakrishnan, Fu, Florence, Finn, Dubey, Driess, Ding, Choromanski, Chen, Chebotar, Carbajal, Brown, Brohan, Arenas, and Han]{rt2}
Brianna Zitkovich, Tianhe Yu, Sichun Xu, Peng Xu, Ted Xiao, Fei Xia, Jialin Wu, Paul Wohlhart, Stefan Welker, Ayzaan Wahid, Quan Vuong, Vincent Vanhoucke, Huong Tran, Radu Soricut, Anikait Singh, Jaspiar Singh, Pierre Sermanet, Pannag~R. Sanketi, Grecia Salazar, Michael~S. Ryoo, Krista Reymann, Kanishka Rao, Karl Pertsch, Igor Mordatch, Henryk Michalewski, Yao Lu, Sergey Levine, Lisa Lee, Tsang-Wei~Edward Lee, Isabel Leal, Yuheng Kuang, Dmitry Kalashnikov, Ryan Julian, Nikhil~J. Joshi, Alex Irpan, Brian Ichter, Jasmine Hsu, Alexander Herzog, Karol Hausman, Keerthana Gopalakrishnan, Chuyuan Fu, Pete Florence, Chelsea Finn, Kumar~Avinava Dubey, Danny Driess, Tianli Ding, Krzysztof~Marcin Choromanski, Xi~Chen, Yevgen Chebotar, Justice Carbajal, Noah Brown, Anthony Brohan, Montserrat~Gonzalez Arenas, and Kehang Han.
\newblock Rt-2: Vision-language-action models transfer web knowledge to robotic control.
\newblock In Jie Tan, Marc Toussaint, and Kourosh Darvish, editors, \emph{Proceedings of The 7th Conference on Robot Learning}, volume 229 of \emph{Proceedings of Machine Learning Research}, pages 2165--2183. PMLR, 06--09 Nov 2023.

\bibitem[Cheng et~al.(2025)Cheng, Ji, Yang, Gongye, Zou, Kautz, B{\i}y{\i}k, Yin, Liu, and Wang]{navila}
An-Chieh Cheng, Yandong Ji, Zhaojing Yang, Zaitian Gongye, Xueyan Zou, Jan Kautz, Erdem B{\i}y{\i}k, Hongxu Yin, Sifei Liu, and Xiaolong Wang.
\newblock Navila: Legged robot vision-language-action model for navigation.
\newblock In \emph{RSS}, 2025.

\bibitem[Driess et~al.(2025)Driess, Springenberg, Ichter, Yu, Li-Bell, Pertsch, Ren, Walke, Vuong, Shi, et~al.]{pi0.5_ki}
Danny Driess, Jost~Tobias Springenberg, Brian Ichter, Lili Yu, Adrian Li-Bell, Karl Pertsch, Allen~Z Ren, Homer Walke, Quan Vuong, Lucy~Xiaoyang Shi, et~al.
\newblock Knowledge insulating vision-language-action models: Train fast, run fast, generalize better.
\newblock \emph{arXiv preprint arXiv:2505.23705}, 2025.

\bibitem[Ho et~al.(2020)Ho, Jain, and Abbeel]{ddpm}
Jonathan Ho, Ajay Jain, and Pieter Abbeel.
\newblock Denoising diffusion probabilistic models.
\newblock In H.~Larochelle, M.~Ranzato, R.~Hadsell, M.F. Balcan, and H.~Lin, editors, \emph{Advances in Neural Information Processing Systems}, volume~33, pages 6840--6851. Curran Associates, Inc., 2020.
\newblock URL \url{https://proceedings.neurips.cc/paper_files/paper/2020/file/4c5bcfec8584af0d967f1ab10179ca4b-Paper.pdf}.

\bibitem[Song et~al.(2021)Song, Meng, and Ermon]{ddim}
Jiaming Song, Chenlin Meng, and Stefano Ermon.
\newblock Denoising diffusion implicit models.
\newblock In \emph{International Conference on Learning Representations}, 2021.
\newblock URL \url{https://openreview.net/forum?id=St1giarCHLP}.

\bibitem[Chi et~al.(2024)Chi, Xu, Feng, Cousineau, Du, Burchfiel, Tedrake, and Song]{dp}
Cheng Chi, Zhenjia Xu, Siyuan Feng, Eric Cousineau, Yilun Du, Benjamin Burchfiel, Russ Tedrake, and Shuran Song.
\newblock Diffusion policy: Visuomotor policy learning via action diffusion.
\newblock \emph{The International Journal of Robotics Research}, 2024.

\bibitem[Liu et~al.(2025)Liu, Wu, Li, Tan, Chen, Wang, Xu, Su, and Zhu]{rdt}
Songming Liu, Lingxuan Wu, Bangguo Li, Hengkai Tan, Huayu Chen, Zhengyi Wang, Ke~Xu, Hang Su, and Jun Zhu.
\newblock {RDT}-1b: a diffusion foundation model for bimanual manipulation.
\newblock In \emph{The Thirteenth International Conference on Learning Representations}, 2025.
\newblock URL \url{https://openreview.net/forum?id=yAzN4tz7oI}.

\bibitem[Barreiros et~al.(2025{\natexlab{a}})Barreiros, Beaulieu, Bhat, Cory, Cousineau, Dai, Fang, Hashimoto, Irshad, Itkina, et~al.]{barreiros2025careful}
Jose Barreiros, Andrew Beaulieu, Aditya Bhat, Rick Cory, Eric Cousineau, Hongkai Dai, Ching-Hsin Fang, Kunimatsu Hashimoto, Muhammad~Zubair Irshad, Masha Itkina, et~al.
\newblock A careful examination of large behavior models for multitask dexterous manipulation.
\newblock \emph{arXiv preprint arXiv:2507.05331}, 2025{\natexlab{a}}.

\bibitem[Intelligence et~al.(2025)Intelligence, Black, Brown, Darpinian, Dhabalia, Driess, Esmail, Equi, Finn, Fusai, et~al.]{pi0.5}
Physical Intelligence, Kevin Black, Noah Brown, James Darpinian, Karan Dhabalia, Danny Driess, Adnan Esmail, Michael Equi, Chelsea Finn, Niccolo Fusai, et~al.
\newblock $\pi$ 0.5: a vision-language-action model with open-world generalization.
\newblock \emph{arXiv preprint arXiv:2504.16054}, 2025.

\bibitem[Liang et~al.()Liang, YU, Luo, Iyer, Dong, Zhou, Ghosh, Lewis, Yih, Zettlemoyer, et~al.]{mot}
Weixin Liang, LILI YU, Liang Luo, Srini Iyer, Ning Dong, Chunting Zhou, Gargi Ghosh, Mike Lewis, Wen-tau Yih, Luke Zettlemoyer, et~al.
\newblock Mixture-of-transformers: A sparse and scalable architecture for multi-modal foundation models.
\newblock In \emph{ICLR 2025 Workshop on World Models: Understanding, Modelling and Scaling}.

\bibitem[Gu et~al.(2023)Gu, Xiang, Li, Ling, Liu, Mu, Tang, Tao, Wei, Yao, Yuan, Xie, Huang, Chen, and Su]{maniskill}
Jiayuan Gu, Fanbo Xiang, Xuanlin Li, Zhan Ling, Xiqiang Liu, Tongzhou Mu, Yihe Tang, Stone Tao, Xinyue Wei, Yunchao Yao, Xiaodi Yuan, Pengwei Xie, Zhiao Huang, Rui Chen, and Hao Su.
\newblock Maniskill2: A unified benchmark for generalizable manipulation skills.
\newblock In \emph{International Conference on Learning Representations}, 2023.

\bibitem[Khazatsky et~al.()Khazatsky, Pertsch, Nair, Balakrishna, Dasari, Karamcheti, Nasiriany, Srirama, Chen, Ellis, et~al.]{droid}
Alexander Khazatsky, Karl Pertsch, Suraj Nair, Ashwin Balakrishna, Sudeep Dasari, Siddharth Karamcheti, Soroush Nasiriany, Mohan~Kumar Srirama, Lawrence~Yunliang Chen, Kirsty Ellis, et~al.
\newblock Droid: A large-scale in-the-wild robot manipulation dataset.
\newblock In \emph{RSS 2024 Workshop: Data Generation for Robotics}.

\bibitem[O’Neill et~al.(2024)O’Neill, Rehman, Maddukuri, Gupta, Padalkar, Lee, Pooley, Gupta, Mandlekar, Jain, et~al.]{xemb}
Abby O’Neill, Abdul Rehman, Abhiram Maddukuri, Abhishek Gupta, Abhishek Padalkar, Abraham Lee, Acorn Pooley, Agrim Gupta, Ajay Mandlekar, Ajinkya Jain, et~al.
\newblock Open x-embodiment: Robotic learning datasets and rt-x models: Open x-embodiment collaboration 0.
\newblock In \emph{2024 IEEE International Conference on Robotics and Automation (ICRA)}, pages 6892--6903. IEEE, 2024.

\bibitem[Walke et~al.(2023)Walke, Black, Zhao, Vuong, Zheng, Hansen-Estruch, He, Myers, Kim, Du, et~al.]{bridge}
Homer~Rich Walke, Kevin Black, Tony~Z Zhao, Quan Vuong, Chongyi Zheng, Philippe Hansen-Estruch, Andre~Wang He, Vivek Myers, Moo~Jin Kim, Max Du, et~al.
\newblock Bridgedata v2: A dataset for robot learning at scale.
\newblock In \emph{Conference on Robot Learning}, pages 1723--1736. PMLR, 2023.

\bibitem[Ji et~al.(2025)Ji, Tan, Shi, Hao, Zhang, Zhang, Wang, Zhao, Mu, An, et~al.]{robobrain}
Yuheng Ji, Huajie Tan, Jiayu Shi, Xiaoshuai Hao, Yuan Zhang, Hengyuan Zhang, Pengwei Wang, Mengdi Zhao, Yao Mu, Pengju An, et~al.
\newblock Robobrain: A unified brain model for robotic manipulation from abstract to concrete.
\newblock In \emph{Proceedings of the Computer Vision and Pattern Recognition Conference}, pages 1724--1734, 2025.

\bibitem[Chen et~al.(2025{\natexlab{a}})Chen, Xie, Ma, Sanketi, and Goldberg]{robo2vlm}
Kaiyuan Chen, Shuangyu Xie, Zehan Ma, Pannag~R Sanketi, and Ken Goldberg.
\newblock Robo2vlm: Visual question answering from large-scale in-the-wild robot manipulation datasets.
\newblock \emph{arXiv preprint arXiv:2505.15517}, 2025{\natexlab{a}}.

\bibitem[Li et~al.(2025{\natexlab{a}})Li, Mata, Park, Kahatapitiya, Jang, Shang, Ranasinghe, Burgert, Cai, Lee, and Ryoo]{llara}
Xiang Li, Cristina Mata, Jongwoo Park, Kumara Kahatapitiya, Yoo~Sung Jang, Jinghuan Shang, Kanchana Ranasinghe, Ryan Burgert, Mu~Cai, Yong~Jae Lee, and Michael~S. Ryoo.
\newblock Llara: Supercharging robot learning data for vision-language policy.
\newblock In \emph{International Conference on Learning Representations}, 2025{\natexlab{a}}.

\bibitem[Team et~al.(2025)Team, Abeyruwan, Ainslie, Alayrac, Arenas, Armstrong, Balakrishna, Baruch, Bauza, Blokzijl, et~al.]{geminirobotics}
Gemini~Robotics Team, Saminda Abeyruwan, Joshua Ainslie, Jean-Baptiste Alayrac, Montserrat~Gonzalez Arenas, Travis Armstrong, Ashwin Balakrishna, Robert Baruch, Maria Bauza, Michiel Blokzijl, et~al.
\newblock Gemini robotics: Bringing ai into the physical world.
\newblock \emph{arXiv preprint arXiv:2503.20020}, 2025.

\bibitem[Xue et~al.(2025)Xue, Ren, Chen, Zhang, Fang, Gu, Xu, and Lu]{rdp}
Han Xue, Jieji Ren, Wendi Chen, Gu~Zhang, Yuan Fang, Guoying Gu, Huazhe Xu, and Cewu Lu.
\newblock Reactive diffusion policy: Slow-fast visual-tactile policy learning for contact-rich manipulation.
\newblock In \emph{Proceedings of Robotics: Science and Systems (RSS)}, 2025.

\bibitem[Qi et~al.(2025)Qi, Yin, and Yang]{qi2024control}
Han Qi, Haocheng Yin, and Heng Yang.
\newblock Control-oriented clustering of visual latent representation.
\newblock In \emph{International Conference on Learning Representations}, 2025.

\bibitem[Wen et~al.(2025)Wen, Zhu, Li, Zhu, Tang, Wu, Xu, Liu, Cheng, Shen, et~al.]{tinyvla}
Junjie Wen, Yichen Zhu, Jinming Li, Minjie Zhu, Zhibin Tang, Kun Wu, Zhiyuan Xu, Ning Liu, Ran Cheng, Chaomin Shen, et~al.
\newblock Tinyvla: Towards fast, data-efficient vision-language-action models for robotic manipulation.
\newblock \emph{IEEE Robotics and Automation Letters}, 2025.

\bibitem[Deng et~al.(2025{\natexlab{a}})Deng, Yan, Wei, Ma, Yang, Chen, Zhang, Yang, Zhang, Cui, et~al.]{graspvla}
Shengliang Deng, Mi~Yan, Songlin Wei, Haixin Ma, Yuxin Yang, Jiayi Chen, Zhiqi Zhang, Taoyu Yang, Xuheng Zhang, Heming Cui, et~al.
\newblock Graspvla: a grasping foundation model pre-trained on billion-scale synthetic action data.
\newblock \emph{arXiv preprint arXiv:2505.03233}, 2025{\natexlab{a}}.

\bibitem[Chen et~al.(2025{\natexlab{b}})Chen, Wei, Zhang, Zhang, Wang, Guo, Yang, Wang, Xiao, Zhao, Chen, and Bian]{villa-x}
Xiaoyu Chen, Hangxing Wei, Pushi Zhang, Chuheng Zhang, Kaixin Wang, Yanjiang Guo, Rushuai Yang, Yucen Wang, Xinquan Xiao, Li~Zhao, Jianyu Chen, and Jiang Bian.
\newblock villa-x: Enhancing latent action modeling in vision-language-action models.
\newblock \emph{arXiv preprint arXiv: 2507.23682}, 2025{\natexlab{b}}.

\bibitem[Bjorck et~al.(2025)Bjorck, Casta{\~n}eda, Cherniadev, Da, Ding, Fan, Fang, Fox, Hu, Huang, et~al.]{groot}
Johan Bjorck, Fernando Casta{\~n}eda, Nikita Cherniadev, Xingye Da, Runyu Ding, Linxi Fan, Yu~Fang, Dieter Fox, Fengyuan Hu, Spencer Huang, et~al.
\newblock Gr00t n1: An open foundation model for generalist humanoid robots.
\newblock \emph{arXiv preprint arXiv:2503.14734}, 2025.

\bibitem[Barreiros et~al.(2025{\natexlab{b}})Barreiros, Beaulieu, Bhat, Cory, Cousineau, Dai, Fang, Hashimoto, Irshad, Itkina, et~al.]{lbm}
Jose Barreiros, Andrew Beaulieu, Aditya Bhat, Rick Cory, Eric Cousineau, Hongkai Dai, Ching-Hsin Fang, Kunimatsu Hashimoto, Muhammad~Zubair Irshad, Masha Itkina, et~al.
\newblock A careful examination of large behavior models for multitask dexterous manipulation.
\newblock \emph{arXiv preprint arXiv:2507.05331}, 2025{\natexlab{b}}.

\bibitem[Deng et~al.(2025{\natexlab{b}})Deng, Zhu, Li, Gou, Li, Wang, Zhong, Yu, Nie, Song, et~al.]{bagel}
Chaorui Deng, Deyao Zhu, Kunchang Li, Chenhui Gou, Feng Li, Zeyu Wang, Shu Zhong, Weihao Yu, Xiaonan Nie, Ziang Song, et~al.
\newblock Emerging properties in unified multimodal pretraining.
\newblock \emph{arXiv preprint arXiv:2505.14683}, 2025{\natexlab{b}}.

\bibitem[Tschannen et~al.(2025)Tschannen, Gritsenko, Wang, Naeem, Alabdulmohsin, Parthasarathy, Evans, Beyer, Xia, Mustafa, et~al.]{tschannen2025siglip}
Michael Tschannen, Alexey Gritsenko, Xiao Wang, Muhammad~Ferjad Naeem, Ibrahim Alabdulmohsin, Nikhil Parthasarathy, Talfan Evans, Lucas Beyer, Ye~Xia, Basil Mustafa, et~al.
\newblock Siglip 2: Multilingual vision-language encoders with improved semantic understanding, localization, and dense features.
\newblock \emph{arXiv preprint arXiv:2502.14786}, 2025.

\bibitem[Hui et~al.(2024)Hui, Yang, Cui, Yang, Liu, Zhang, Liu, Zhang, Yu, Lu, et~al.]{hui2024qwen2}
Binyuan Hui, Jian Yang, Zeyu Cui, Jiaxi Yang, Dayiheng Liu, Lei Zhang, Tianyu Liu, Jiajun Zhang, Bowen Yu, Keming Lu, et~al.
\newblock Qwen2. 5-coder technical report.
\newblock \emph{arXiv preprint arXiv:2409.12186}, 2024.

\bibitem[Zhang and Sennrich(2019)]{rmsnorm}
Biao Zhang and Rico Sennrich.
\newblock Root mean square layer normalization.
\newblock \emph{Advances in neural information processing systems}, 32, 2019.

\bibitem[Su et~al.(2024)Su, Ahmed, Lu, Pan, Bo, and Liu]{rope}
Jianlin Su, Murtadha Ahmed, Yu~Lu, Shengfeng Pan, Wen Bo, and Yunfeng Liu.
\newblock Roformer: Enhanced transformer with rotary position embedding.
\newblock \emph{Neurocomputing}, 568:\penalty0 127063, 2024.

\bibitem[Li et~al.(2024)Li, Zhang, Guo, Zhang, Li, Zhang, Zhang, Zhang, Li, Liu, et~al.]{llavaov}
Bo~Li, Yuanhan Zhang, Dong Guo, Renrui Zhang, Feng Li, Hao Zhang, Kaichen Zhang, Peiyuan Zhang, Yanwei Li, Ziwei Liu, et~al.
\newblock Llava-onevision: Easy visual task transfer.
\newblock \emph{arXiv preprint arXiv:2408.03326}, 2024.

\bibitem[Papineni et~al.(2002)Papineni, Roukos, Ward, and Zhu]{bleu}
Kishore Papineni, Salim Roukos, Todd Ward, and Wei-Jing Zhu.
\newblock Bleu: a method for automatic evaluation of machine translation.
\newblock In \emph{Proceedings of the 40th annual meeting of the Association for Computational Linguistics}, pages 311--318, 2002.

\bibitem[Banerjee and Lavie(2005)]{meteor}
Satanjeev Banerjee and Alon Lavie.
\newblock Meteor: An automatic metric for mt evaluation with improved correlation with human judgments.
\newblock In \emph{Proceedings of the acl workshop on intrinsic and extrinsic evaluation measures for machine translation and/or summarization}, pages 65--72, 2005.

\bibitem[Vedantam et~al.(2015)Vedantam, Lawrence~Zitnick, and Parikh]{cider}
Ramakrishna Vedantam, C~Lawrence~Zitnick, and Devi Parikh.
\newblock Cider: Consensus-based image description evaluation.
\newblock In \emph{Proceedings of the IEEE conference on computer vision and pattern recognition}, pages 4566--4575, 2015.

\bibitem[Li et~al.(2025{\natexlab{b}})Li, Wu, Huang, Cheang, Wang, and Kong]{grmg}
Peiyan Li, Hongtao Wu, Yan Huang, Chilam Cheang, Liang Wang, and Tao Kong.
\newblock Gr-mg: Leveraging partially-annotated data via multi-modal goal-conditioned policy.
\newblock \emph{IEEE Robotics and Automation Letters}, 2025{\natexlab{b}}.

\bibitem[Black et~al.(2024)Black, Brown, Driess, Esmail, Equi, Finn, Fusai, Groom, Hausman, Ichter, et~al.]{pi0}
Kevin Black, Noah Brown, Danny Driess, Adnan Esmail, Michael Equi, Chelsea Finn, Niccolo Fusai, Lachy Groom, Karol Hausman, Brian Ichter, et~al.
\newblock $pi\_0 $: A vision-language-action flow model for general robot control.
\newblock \emph{arXiv preprint arXiv:2410.24164}, 2024.

\bibitem[Kahou et~al.(2017)Kahou, Michalski, Atkinson, K{\'a}d{\'a}r, Trischler, and Bengio]{figureqa}
Samira~Ebrahimi Kahou, Vincent Michalski, Adam Atkinson, {\'A}kos K{\'a}d{\'a}r, Adam Trischler, and Yoshua Bengio.
\newblock Figureqa: An annotated figure dataset for visual reasoning.
\newblock \emph{arXiv preprint arXiv:1710.07300}, 2017.

\bibitem[Johnson et~al.(2017)Johnson, Hariharan, Van Der~Maaten, Fei-Fei, Lawrence~Zitnick, and Girshick]{clevr}
Justin Johnson, Bharath Hariharan, Laurens Van Der~Maaten, Li~Fei-Fei, C~Lawrence~Zitnick, and Ross Girshick.
\newblock Clevr: A diagnostic dataset for compositional language and elementary visual reasoning.
\newblock In \emph{Proceedings of the IEEE conference on computer vision and pattern recognition}, pages 2901--2910, 2017.

\bibitem[Saikh et~al.(2022)Saikh, Ghosal, Mittal, Ekbal, and Bhattacharyya]{scienceqa}
Tanik Saikh, Tirthankar Ghosal, Amish Mittal, Asif Ekbal, and Pushpak Bhattacharyya.
\newblock Scienceqa: A novel resource for question answering on scholarly articles.
\newblock \emph{International Journal on Digital Libraries}, 23\penalty0 (3):\penalty0 289--301, 2022.

\bibitem[Kiela et~al.(2020)Kiela, Firooz, Mohan, Goswami, Singh, Ringshia, and Testuggine]{hatefulmemes}
Douwe Kiela, Hamed Firooz, Aravind Mohan, Vedanuj Goswami, Amanpreet Singh, Pratik Ringshia, and Davide Testuggine.
\newblock The hateful memes challenge: Detecting hate speech in multimodal memes.
\newblock 2020.

\bibitem[Zhu et~al.(2016)Zhu, Groth, Bernstein, and Fei-Fei]{visual7w}
Yuke Zhu, Oliver Groth, Michael Bernstein, and Li~Fei-Fei.
\newblock Visual7w: Grounded question answering in images.
\newblock In \emph{Proceedings of the IEEE conference on computer vision and pattern recognition}, pages 4995--5004, 2016.

\bibitem[Liu et~al.(2023)Liu, Emerson, and Collier]{vsr}
Fangyu Liu, Guy Emerson, and Nigel Collier.
\newblock Visual spatial reasoning.
\newblock \emph{Transactions of the Association for Computational Linguistics}, 11:\penalty0 635--651, 2023.

\bibitem[Acharya et~al.(2019)Acharya, Kafle, and Kanan]{tallyqa}
Manoj Acharya, Kushal Kafle, and Christopher Kanan.
\newblock Tallyqa: Answering complex counting questions.
\newblock In \emph{Proceedings of the AAAI conference on artificial intelligence}, volume~33, pages 8076--8084, 2019.

\bibitem[Yue et~al.(2024)Yue, Ni, Zhang, Zheng, Liu, Zhang, Stevens, Jiang, Ren, Sun, et~al.]{mmmu}
Xiang Yue, Yuansheng Ni, Kai Zhang, Tianyu Zheng, Ruoqi Liu, Ge~Zhang, Samuel Stevens, Dongfu Jiang, Weiming Ren, Yuxuan Sun, et~al.
\newblock Mmmu: A massive multi-discipline multimodal understanding and reasoning benchmark for expert agi.
\newblock In \emph{Proceedings of the IEEE/CVF Conference on Computer Vision and Pattern Recognition}, pages 9556--9567, 2024.

\end{thebibliography}

\clearpage

\appendix
\section{Appendix}\label{sec:appendix}

\subsection{Reasoning Latency.} \label{latency}
Low reasoning latency is one of the advantage of the MoTVLA as it is able to perform fast reasoning with domain expert. We have thoroughly benchmarked the reasoning frequency of MoTVLA with two different sizes comparing with the $\pi$0.5\_KI, which has the similar motion reasoning capability but still within next token prediction paradigm, as well as several open-sourced VLMs. All models are evaluated with the same visual and textual inputs, with the maximum generation length restricted to 16 tokens and computations performed in bfloat16 precision. We measured the reasoning latency with 50 iterations for the same example to avoid the influence of cold start. As shown in Table \ref{reasoning-latency}, MoTVLA-14B significantly outperforms the 3B and 7B baselines on both H100 and A6000 GPUs. Moreover, when scaled down to 1B parameters, MoTVLA achieves a reasoning frequency nearly four times higher than that of LLaVA-OV-0.5B. It is important to note that MoTVLA-0.5B is used only to measure the inference latency of fast reasoning and has not yet been fully integrated into our framework, as pre-training a 0.5B generalist remains highly challenging and requires large-scale datasets. All quantitative and qualitative results reported in this paper are obtained with MoTVLA-14B.

\begin{table*}[h]
\caption{Composition of the dataset employed for supervised fine-tuning.}
\centering
\resizebox{0.5\textwidth}{!}{
\begin{tabular}{c | c | c | c}
\toprule
{\textbf{Model}} & {\textbf{Size}} & \textbf{{H100}} & \textbf{A6000} \\
\midrule
\multirow{1}{*}{$\pi$0.5 KI } &
 3B & $<$ 1Hz & $<$ 1Hz  \\
\multirow{1}{*}{Bagel-VLM } &
7B & 2 Hz & 1 Hz  \\

\multirow{1}{*}{Qwen2.5-VL } &
7B & 3 Hz & 2 Hz  \\
\multirow{1}{*}{LLaVA-OV } &
7B & 3 Hz & 2 Hz  \\
\multirow{1}{*}{LLaVA-OV } &
0.5B & 5 Hz & 4 Hz  \\
\midrule
\multirow{2}{*}{MoTVLA} & 14B& 9 Hz & 4 Hz  \\
 & 1B & 20 Hz & 11 Hz  \\
\bottomrule
\end{tabular}
}
\label{reasoning-latency}
\end{table*}

\clearpage


\subsection{Dataset Composition.}\label{sft-dataset}
Our SFT dataset for training the domain expert consists of 279.6K demonstrations collected in-house, 678K robotic data curated from Robo2VLM \citep{robo2vlm}, and 318K general knowledge samples open-sourced from the internet. The detailed composition of each dataset is summarized in Table \ref{sft-table}.

\begin{table*}[h]
\caption{Composition of the dataset employed for supervised fine-tuning.}
\centering
\resizebox{0.6\textwidth}{!}{
\begin{tabular}{c | c | c }
\toprule
{\textbf{Domain}} & {\textbf{Dataset}} & \textbf{{Size}} \\
\midrule
\multirow{8}{*}{Robotics} & Cube Stacking& 32.2K  \\
 & Peg-in-Hole & 45.3K  \\
 & L-tool Pull & 76.9K   \\
 & Pick \& Place: Eggplant & 11.0K  \\
 & Pick \& Place: Carrot & 10.0K  \\
 & Pick \& Place: Corn & 9.8K\\
 & Table Bussing & 94.4K  \\
 & Robo2VLM$^*$ & 678.0K  \\
\midrule
\multirow{8}{*}{LLaVA-OV VQA} &
 FigureQA & 100.0K  \\
 & CLEVR & 70.0K \\
 & ScienceQA & 5.0K  \\
 & ScienceQA(nona context) & 19.2K  \\
 & Hateful-memes  & 8.5K  \\
 & Visual-7W & 14.4K  \\
 & Visual Spatial Reasoning & 2.2K  \\
 & TallyQA & 98.7K  \\
\bottomrule
\end{tabular}
}
\label{sft-table}
\end{table*}

\subsection{Tasks and Motion Annotation.}\label{annotation}

In this section, we elaborate on the demonstration tasks and corresponding motion annotation.

\paragraph{Object relocation with stability (Cube Stacking).} The robot must pick up a red cube, stack it on a green cube, and release without the stack falling. \emph{Motion decomposition:} (i) “Move toward the red cube.” (ii) “Pick up the red cube." (iii) "Move it to green cube.” (iv) “Stack the red cube on top of the green cube.”

\paragraph{Tight-tolerance insertion (Peg–in–Hole).} The robot must pick up an orange–white peg and insert the orange end into a box with a hole. To increase the difficulty, we manually augment the scene with distractions (a blue solid marker and a red dot marker). to the default setting. \emph{Motion decomposition:} (i) “Move to the location of the peg on the table.” (ii) “Pick up the peg from the table.” (iii) "Move to the box with the hole." (iv) “Insert the orange end of the peg into the box.”

\paragraph{Tool-mediated manipulation beyond reach (L-tool Pull).} The robot must grasp an L-shaped tool and use it to pull a cube that lies beyond the arm’s direct reach. To increase the difficulty, we manually augment the scene with distractions (a green sphere and a white socket). \emph{Motion decomposition:} (i) “Move the robot's end effector to the position of the tool on the table.” (ii) “Pick up the tool from the table.” (iii) “Use the tool to pull the small blue cube that is out of reach.”

\paragraph{Single-stage Pick-and-Place (Vegetables $\rightarrow$ Basket).} The robot must pick up a vegetable and place it into the basket. \textit{Motion decomposition:} (i) ``Move toward the \{\texttt{place\_holder}\}.'' (ii) ``Grasp the \{\texttt{place\_holder}\} (iii) Move it toward the basket.'' (iv) ``Put the \{\texttt{place\_holder}\} in the basket.'' \textit{Note:} \{\texttt{place\_holder}\} corresponds to one of \{corn, eggplant, carrot\}; During the evaluation phase, we introduce three additional tasks containing distractions (a potato and a random cube) in the scene and assess the zero-shot performance.

\begin{figure*}[t]
  \centering
  \includegraphics[width=0.9\textwidth]{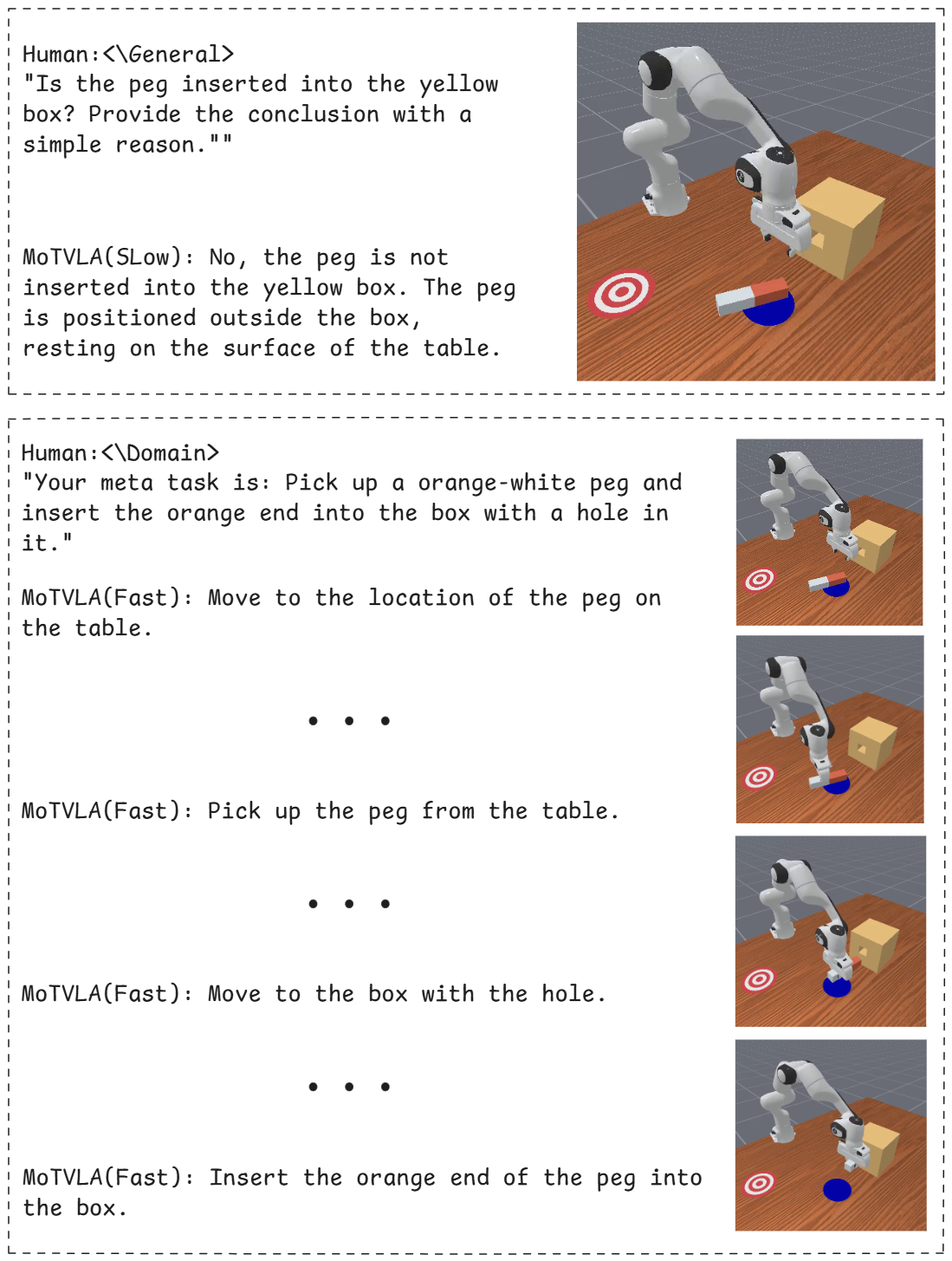}
  \caption{\textbf{A slow and fast reasoning annotation example of MoTVLA in the simulative environment.} }
  \label{example-sim}
\end{figure*}

\begin{figure*}[t]
  \centering
  \includegraphics[width=0.8\textwidth]{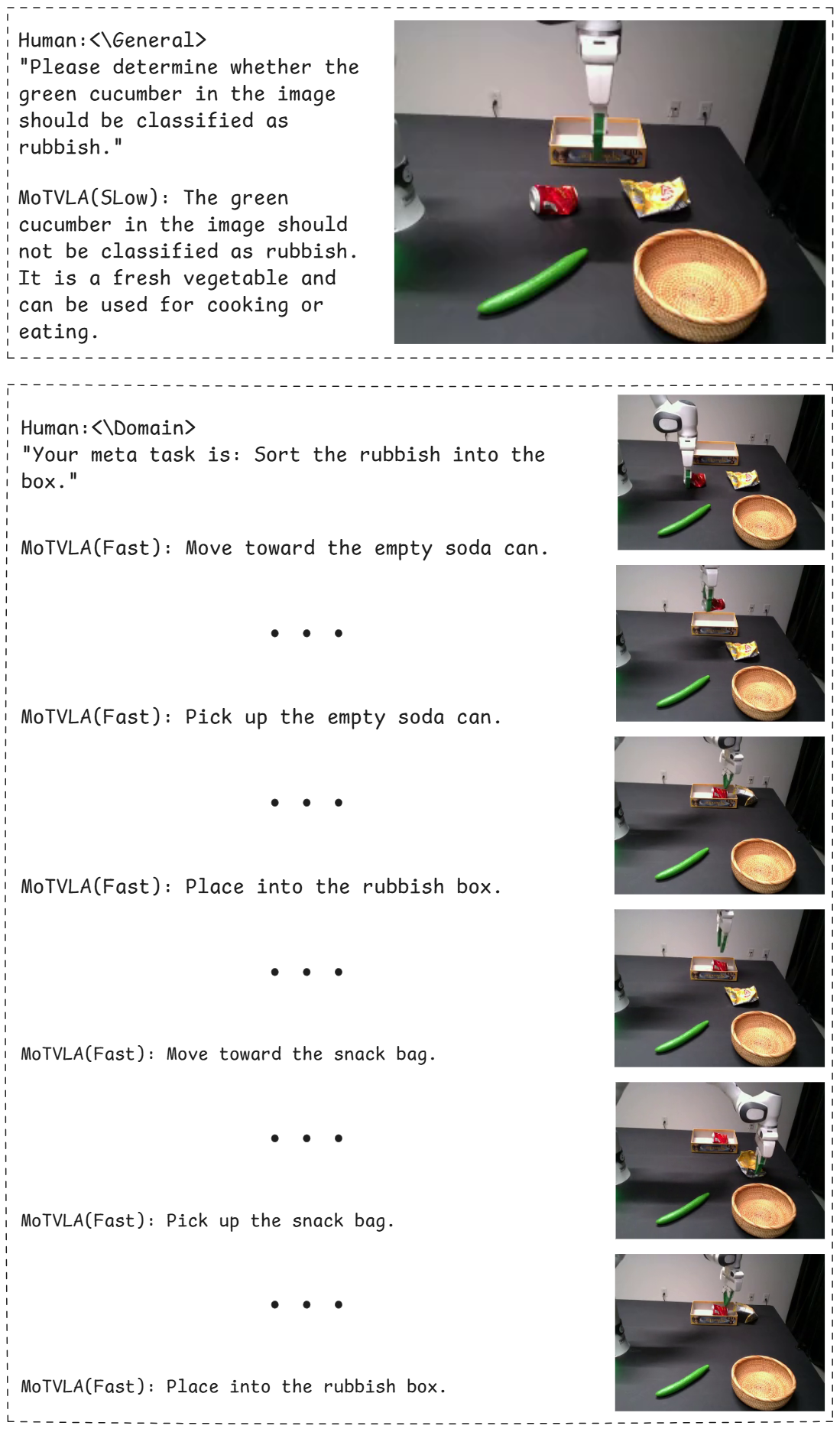}
  \caption{\textbf{A slow and fast reasoning annotation example of MoTVLA in the real-world environment.} }
  \label{example-real}
\end{figure*}

\paragraph{Table Bussing (Ambiguous Instruction).} The scene contains two \emph{rubbish} items (an empty soda can and a snack bag), one \emph{other} item (a cucumber), a yellow box, and a basket. Since the instruction prompt does not explicitly specify which objects are rubbish, the model must interpret the instruction, ground the object categories, and generate a correct motion decomposition and execution order.
\begin{itemize}

    \item \emph{Ambiguous instruction:} ``Sort the rubbish into the box.'' \textit{Motion decomposition:} (i) ``Move toward the empty soda can.'' (ii) ``Pick up the empty soda can."" (iii) Place into the rubbish box.'' (iv) ``Move toward the snack bag.'' (v) ``Pick up the snack bag." (vi) ""Place into the rubbish box.''
    
    \item \emph{Clear instruction:} ``Put the green cucumber into the basket.'' \textit{Motion decomposition:} (i) ``Move toward the cucumber.'' (ii) ``Pick up the cucumber (iii) Place it into the basket.''
\end{itemize}

We collect 300 expert demonstrations for each of the task in simulation, 50 demonstrations for each vegetable pick-and-place task (150 in total), and 100 demonstrations for each table bussing task (200 in total). Each demonstration is segmented into \emph{motion decompositions} with aligned textual descriptions. We illustrate two concrete instances in Fig.\ref{example-sim} and Fig.\ref{example-real}. All baselines are trained on the identical datasets, with language input provided only to methods that initially support text conditioning. Notably, we observe that the motions of \textit{picking up} and \textit{moving toward the destination} are often performed simultaneously by the robot arm, making it difficult to define a clear boundary. Therefore, we merge these two motions into one during training.

\clearpage

\subsection{Domain Expert Reasoning Results.}\label{appendix_derr}
In this section, we present qualitative results on both general and robotic VQA reasoning.

\begin{figure*}[h]
  \centering
  \includegraphics[width=0.8\textwidth,  height=1.1\textwidth]{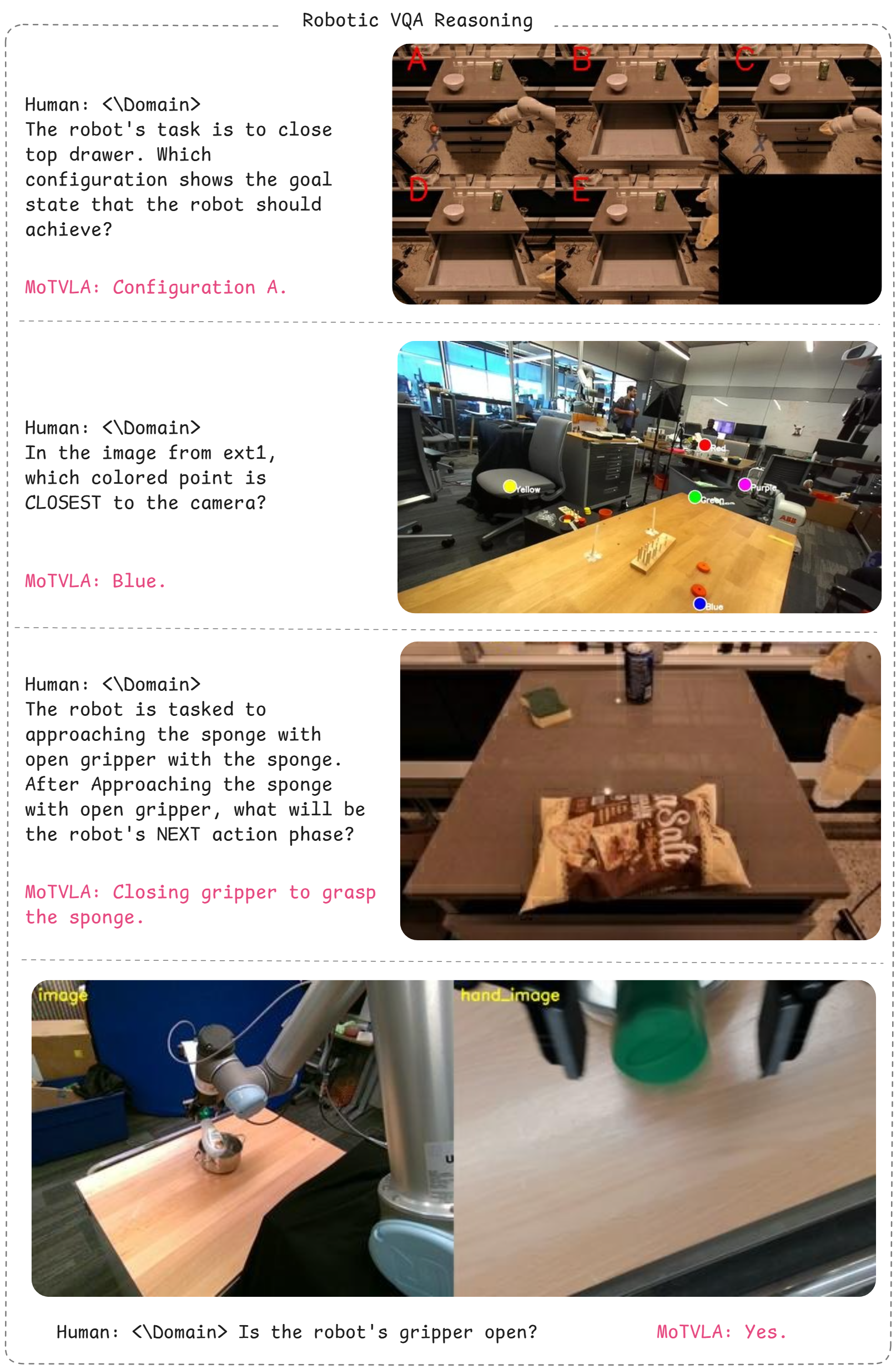}
  \caption{\textbf{Qualitative result of MoTVLA on robotic VQA fast reasoning.} }
  \label{reasoning_1}
\end{figure*}

\begin{figure*}[h]
  \centering
  \includegraphics[width=0.8\textwidth]{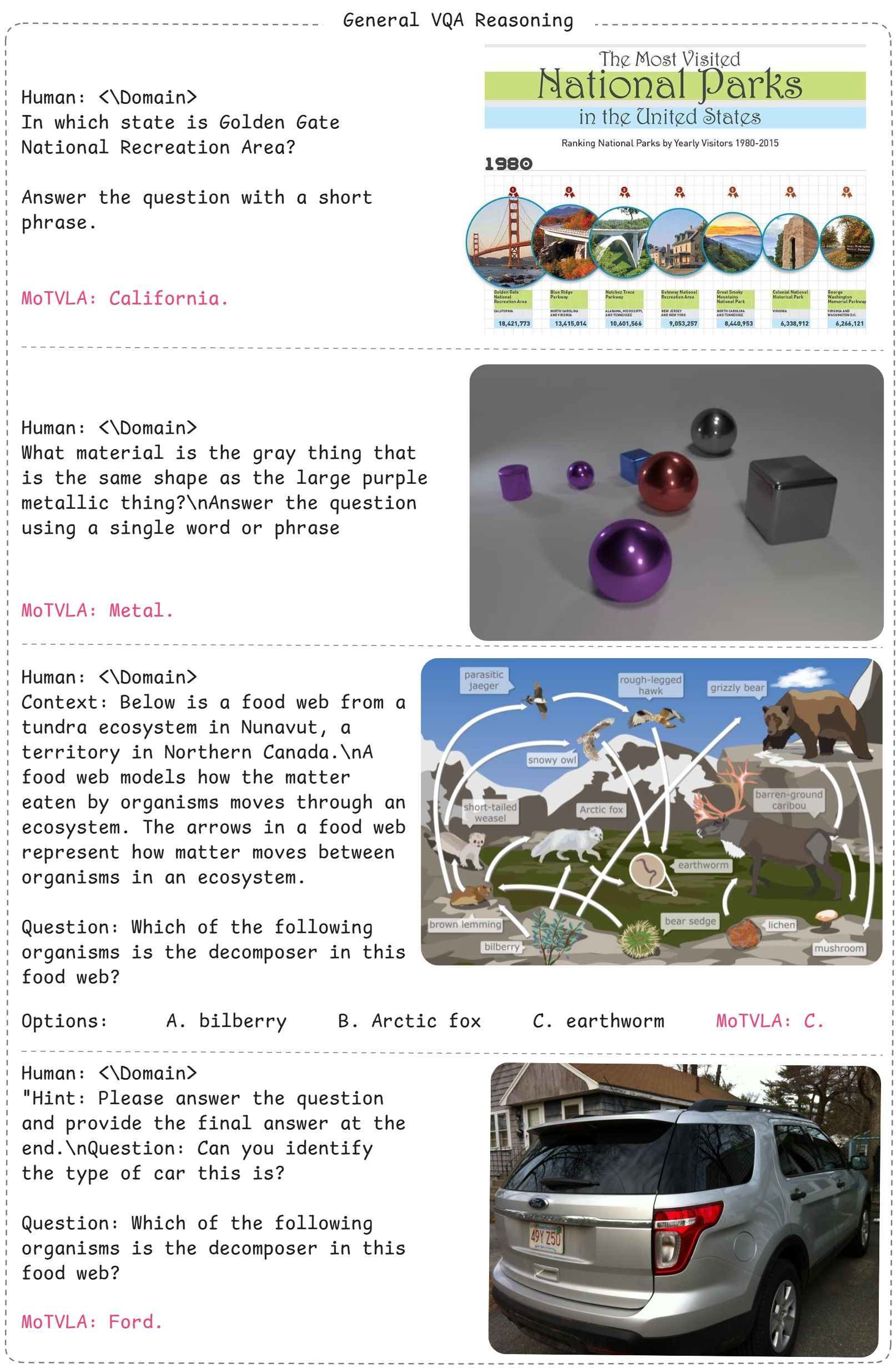}
  \caption{\textbf{Qualitative result of MoTVLA on LLava-OV VQA slow reasoning.} }
  \label{reasoning_2}
\end{figure*}

\clearpage

\subsection{Action Expert Manipulation Experiment Results.}\label{appendix_aemer}

\begin{figure*}[h]
  \centering
  \includegraphics[width=0.8\textwidth,  height=1.1\textwidth]{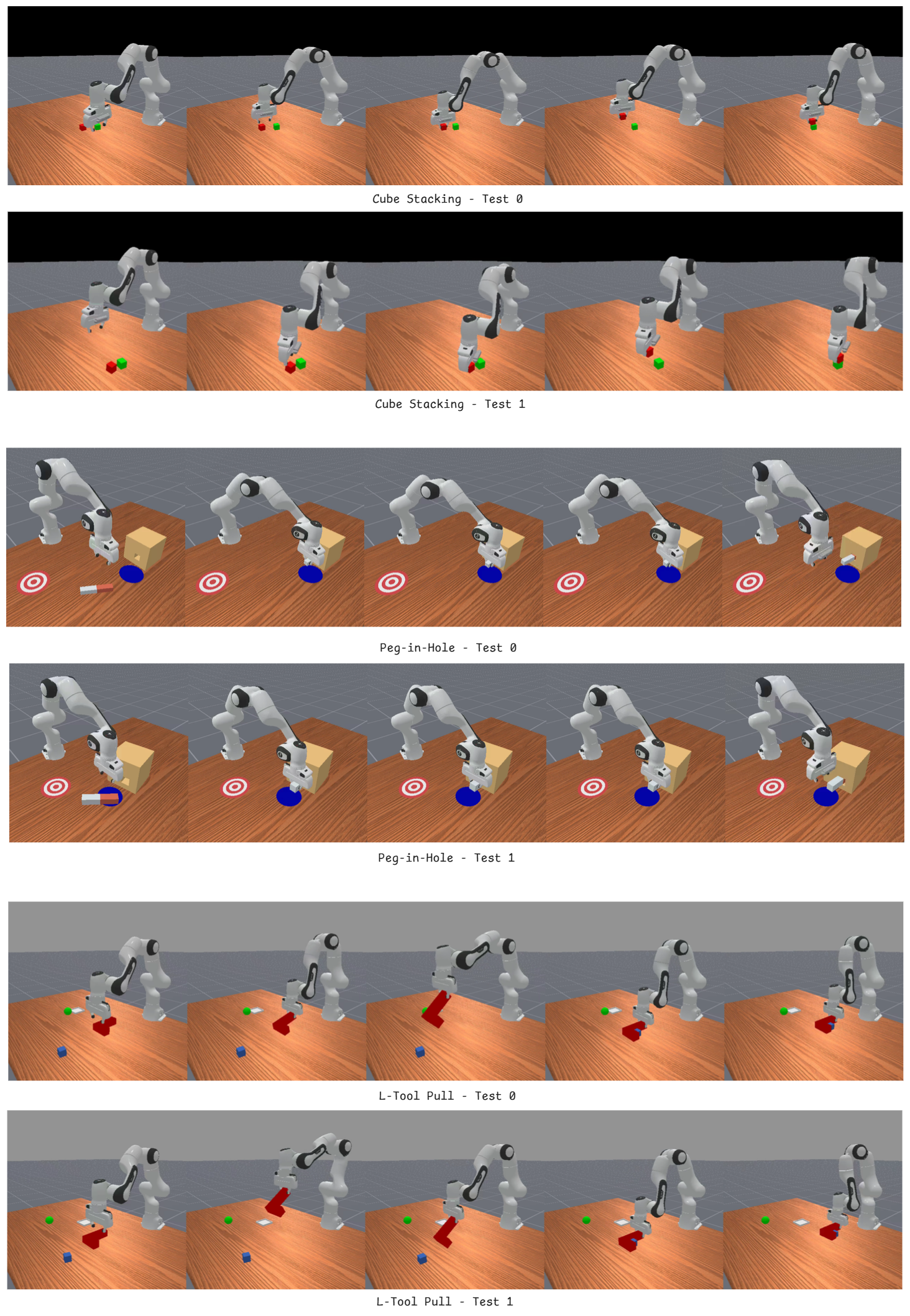}
  \caption{\textbf{Qualitative results of MoTVLA on manipulation tasks in simulation.} All evaluations are conducted with random unseen seeds.}
  \label{rollout_1}
\end{figure*}

\begin{figure*}[t]
  \centering
  \includegraphics[width=0.8\textwidth]{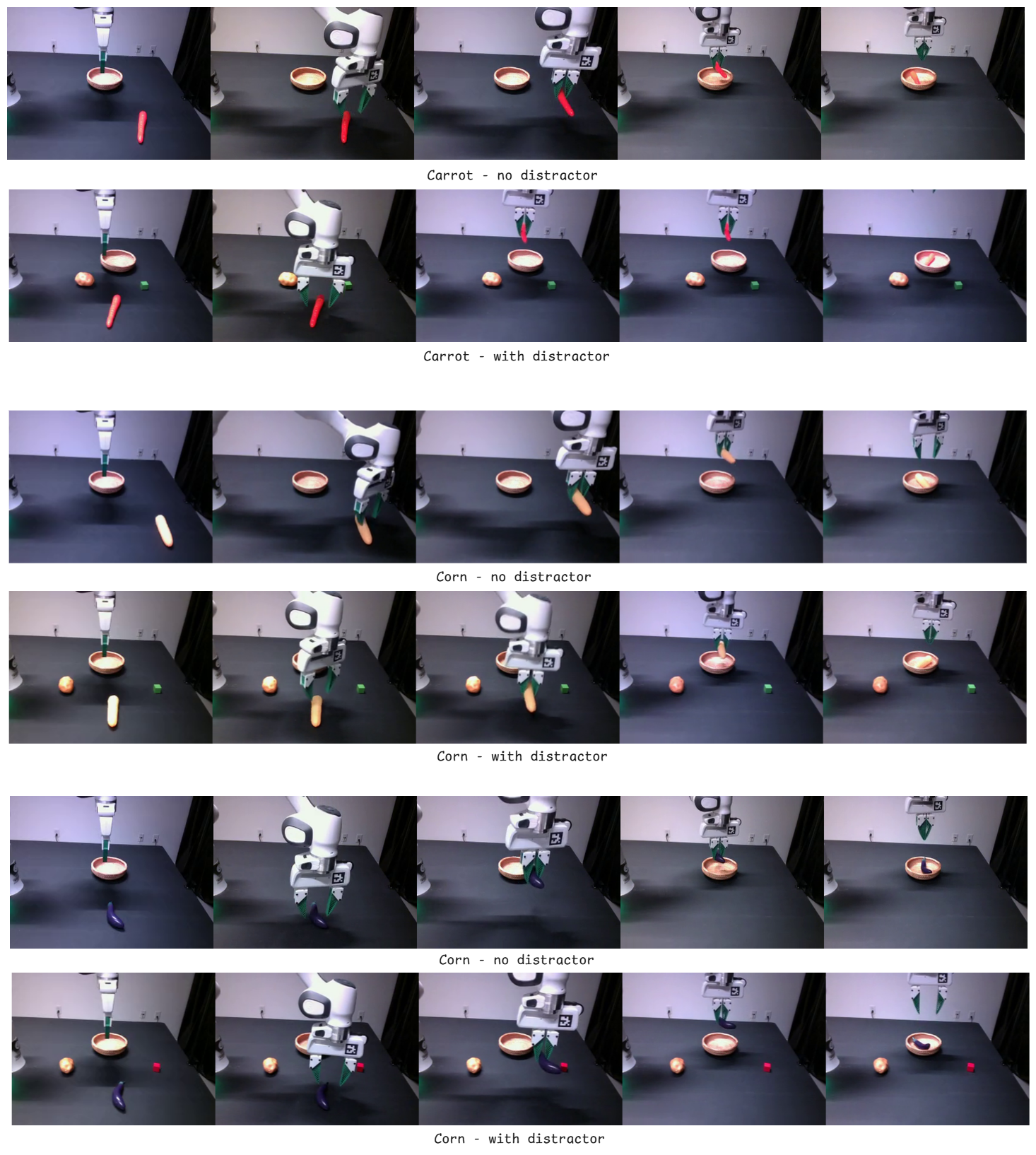}
  \caption{\textbf{Qualitative results of MoTVLA on pick and place tasks in real-world.} All evaluations are conducted with random unseen seeds.}
  \label{rollout_2}
\end{figure*}

\begin{figure*}[t]
  \centering
  \includegraphics[width=0.8\textwidth]{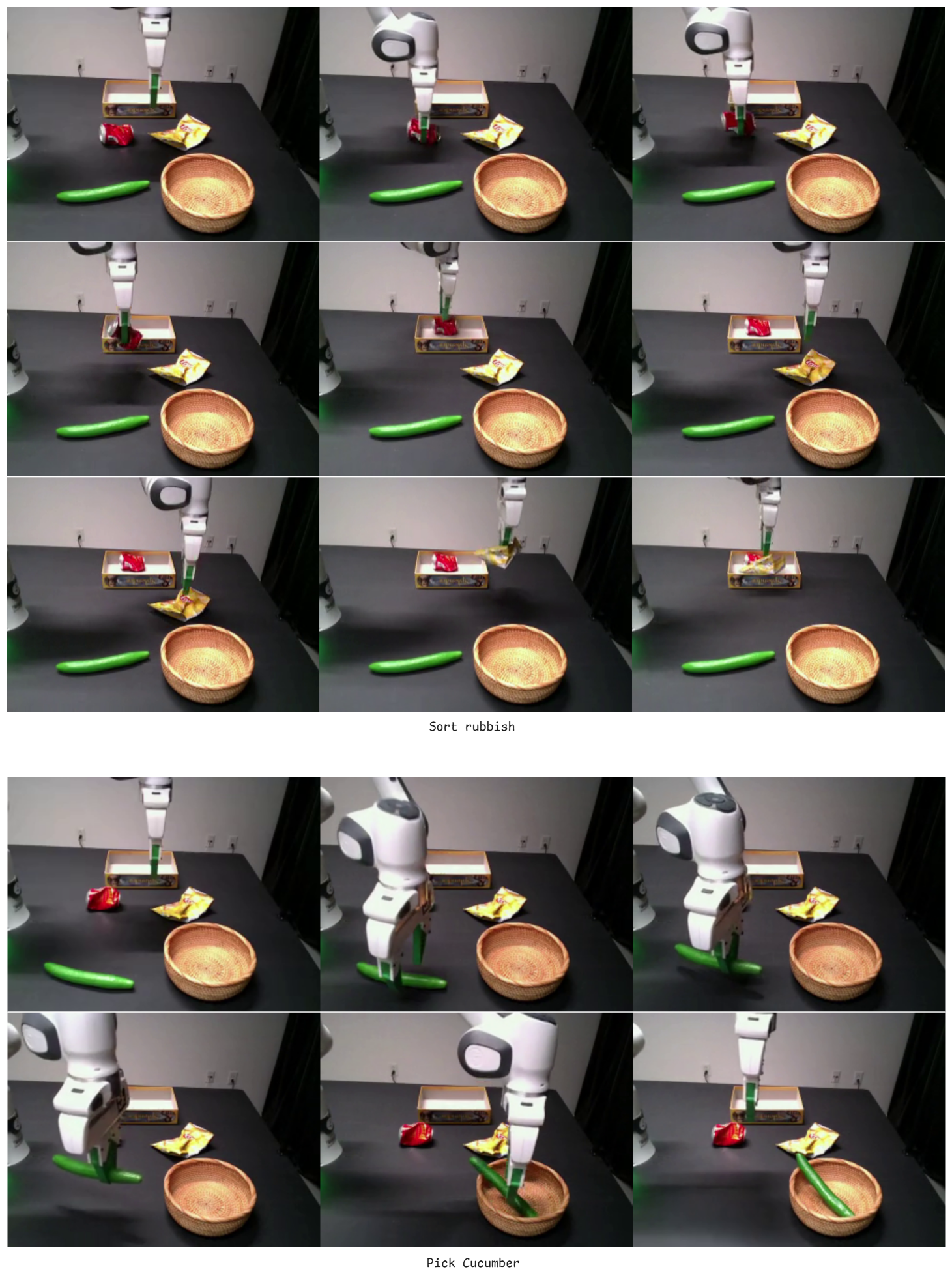}
  \caption{\textbf{Qualitative results of MoTVLA on table bussing tasks in real-world.} All evaluations are conducted with random unseen seeds.}
  \label{rollout_3}
\end{figure*}

\end{document}